\documentclass[lettersize,journal]{IEEEtran}
\usepackage{amsmath,amsfonts}
\usepackage{algorithmic}
\usepackage{algorithm}
\usepackage{array}
\usepackage[caption=false,font=normalsize,labelfont=sf,textfont=sf]{subfig}
\usepackage{textcomp}
\usepackage{stfloats}
\usepackage{url}
\usepackage{verbatim}
\usepackage{graphicx}
\usepackage{cite}
\usepackage{algorithm}
\usepackage{algorithmic}
\usepackage[table,dvipsnames,svgnames]{xcolor}
\usepackage{amsmath}
\usepackage{booktabs}
\usepackage{colortbl}
\usepackage{multirow}
\usepackage{makecell}
\usepackage{pifont}
\newcommand{\cmark}{\ding{51}} 
\newcommand{\incre}[1]{\textcolor{teal!90}{#1}}
\newcommand{\decre}[1]{\textcolor{Bittersweet}{#1}}
\newcommand{\cn}[1]{}
\definecolor{revisionblue}{RGB}{65,105,225}
\DeclareRobustCommand{\revision}[1]{#1} 
\definecolor{tabhighlight}{HTML}{e5e5e5}
\newcommand{\myparagraph}[1]{\noindent\textbf{#1}\,}\usepackage[hidelinks]{hyperref}
\hypersetup{bookmarksopen=true, bookmarksnumbered=true}

\usepackage{enumitem}

\begin{document}

\title{PETR: Prompt Ensembling with Training-free Routing for Vision-Language Models}

\author{Weihan~Cai, Hao~Tan, Xinping~Gao, Shibiao~Xu, and Jun~Wan,~\IEEEmembership{Senior Member,~IEEE}
\thanks{W. Cai, H. Tan, and J. Wan are with the State Key Laboratory of Multimodal Artificial Intelligence Systems, Institute of Automation, Chinese Academy of Sciences, and also with the School of Artificial Intelligence, University of Chinese Academy of Sciences, Beijing, China (e-mail: caiweihan2025@ia.ac.cn; tanhao2023@ia.ac.cn; jun.wan@ia.ac.cn).}%
\thanks{X. Gao is with Purple Mountain Laboratories, Nanjing, China (e-mail: gaoxinping@pmlabs.com.cn).}%
\thanks{S. Xu is with Beijing University of Posts and Telecommunications, Beijing, China (e-mail: shibiaoxu@bupt.edu.cn).}%
}

\markboth{}{}


\maketitle

\begin{abstract}
\revision{
Prompt learning efficiently adapts vision-language models (VLMs) to downstream tasks, but gains on seen classes often come at the expense of generalization to unseen classes.
To address this limitation, we propose \textbf{prompt ensembling with training-free routing (PETR)}, whose key innovation is a carefully designed dual-prompt architecture: two complementary prompts are learned from different data and objectives to emphasize seen-class discrimination and unseen-class generalization, respectively.
}
During training, both prompts are fine-tuned using a shared frozen CLIP backbone, and statistical information is collected from the training set logits. 
At inference time, we determine the similarity of each test sample to seen data, and route the sample to the most appropriate prompt branch.
\revision{
To the best of our knowledge, this is the first prompt tuning framework that performs \textit{training-free adaptive routing} based on statistical similarity.
}
This design provides an interpretable routing signal and avoids common MoE-style routing pathologies, such as router training instability and load imbalance.
Extensive experiments on 11 benchmark datasets demonstrate that our framework consistently outperforms previous methods on both seen and unseen classes, achieving new state-of-the-art results.
\cn{
提示学习能够高效地将视觉—语言模型（VLM）适配到下游任务，但已见类别上的性能提升往往以牺牲未见类别的泛化能力为代价。
为解决这一问题，我们提出 PETR（Prompt Ensembling with Training-free Routing），其核心创新是一种精心设计的双提示架构：两组互补提示通过不同数据和训练目标进行学习，分别侧重已见类判别与未见类泛化。
训练阶段，我们在共享且冻结的 CLIP 主干上分别微调两组 prompt，并从训练集 logits 中收集统计信息。
推理阶段，我们衡量每个测试样本与训练分布之间的相似度，并将样本路由到最合适的 prompt 分支。
据我们所知，这是首个基于统计相似度实现无需训练自适应路由的 prompt 微调框架。这给路由模块带来了良好的可解释性，并且避免路由模块常见的路由训练不稳定问题和负载均衡问题。
在 11 个基准数据集上的大量实验表明，PETR 在已见与未见类别上均稳定优于以往方法，取得新的最先进结果。
}
\end{abstract}

\begin{IEEEkeywords}
Vision-language models, prompt learning, few-shot learning, prompt distillation, model ensembling.
\end{IEEEkeywords}

\section{Introduction}
Large pre-trained vision-language models (VLMs), such as CLIP \cite{CLIP}, have demonstrated remarkable performance on open-vocabulary downstream tasks. These models typically consist of an image encoder and a text encoder, which are jointly trained in a self-supervised manner on large-scale image-text pairs to align image and text features. This alignment enables their application to downstream tasks such as object detection \cite{object-detection-1, object-detection-2}, and image segmentation \cite{segmentation-1, segmentation-2, segmentation-3}.
\cn{
大型预训练视觉-语言模型（VLM），如 CLIP [CLIP]，在开放词汇下游任务中展现了卓越性能。这类模型通常由图像编码器与文本编码器组成，并在大规模图像-文本对上以自监督方式联合训练，从而对齐图像与文本特征，进而支持目标检测 [object-detection-1, object-detection-2]、图像分割 [segmentation-1, segmentation-2, segmentation-3] 等任务。
}

Prompt learning~\cite{CoOp} has emerged as an efficient paradigm for adapting VLMs with minimal trainable parameters by inserting a small number of learnable vectors (soft prompts).
Some studies~\cite{Prograd} suggest that prompt tuning tends to overfit to \emph{seen} classes and generalizes poorly to \emph{unseen} classes.
While prior work improves robustness via regularization or multi-level prompting~\cite{CoCoOp, MaPLe, PromptSRC}, most methods still rely on a single prompt for all test samples, forcing a compromise between seen-class fitting and unseen-class generalization.
\cn{
提示学习 [CoOp] 是一种高效的 VLM 适配范式：通过在输入中插入少量可学习向量（soft prompt），即可在极低参数开销下完成下游适配
一些研究表明 [Prograd]，提示学习倾向于过拟合已见类别，而在未见类别上的泛化能力较差。
尽管已有工作通过正则化或多层提示等策略提升鲁棒性 [CoCoOp, MaPLe, PromptSRC]，多数方法仍使用单一提示处理所有测试样本，使同一组参数不得不在“拟合已见训练分布”与“保持未见泛化”之间做折中。
}

\begin{figure}[t]
    \centering
    \includegraphics[width=\columnwidth]{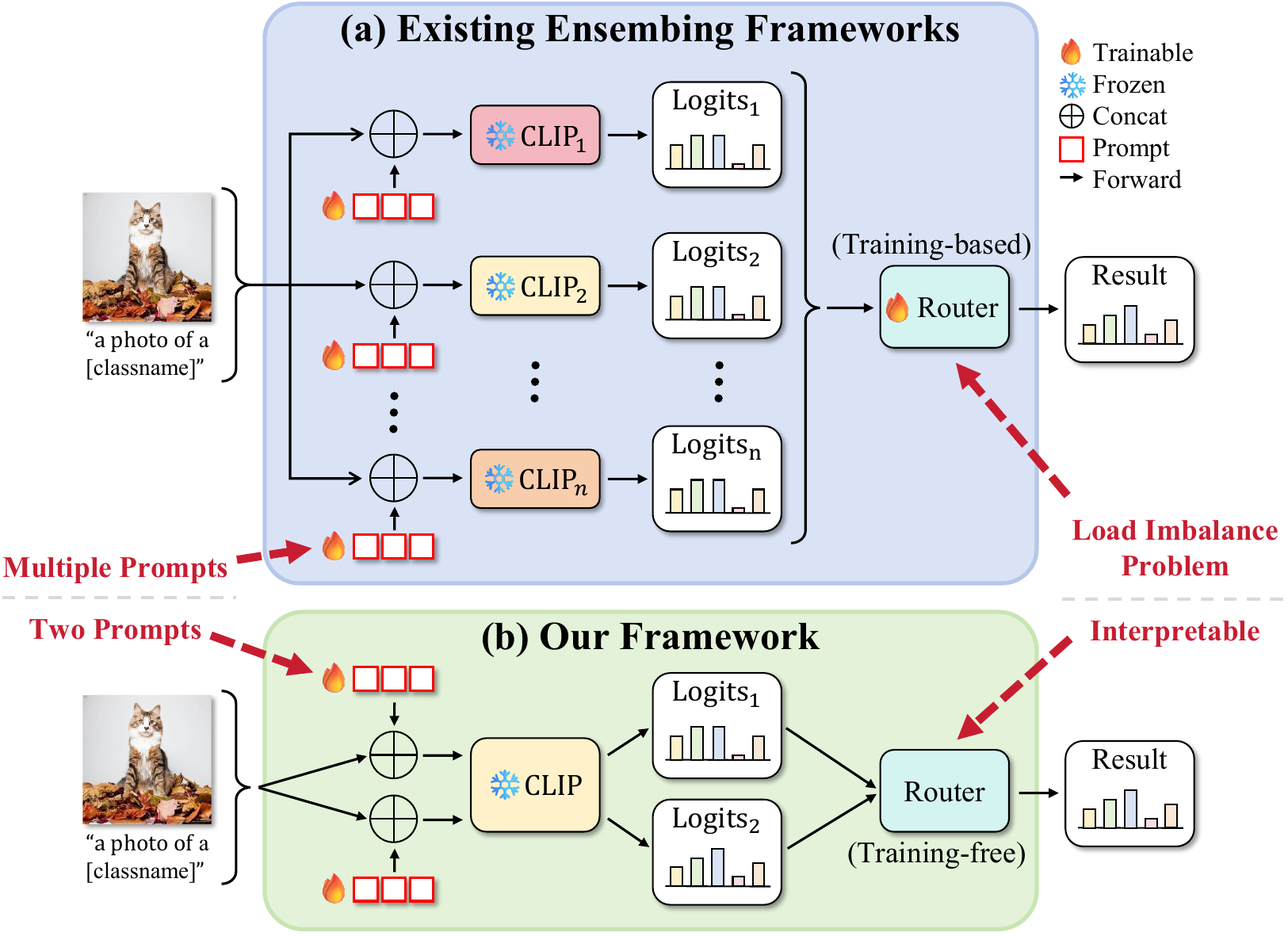}
    \caption{Compare our framework with previous approaches that apply model ensembling to prompt learning. (a) Existing methods ensemble multiple prompt-tuned models with a trainable router. (b) Our framework ensembles two soft prompts with a training-free router, where each prompt is specifically designed to focus on either seen class performance or generalization to unseen classes, respectively.}
    \cn{
    框架对比。（a）现有方法集成多个 prompt 微调模型，并使用可训练的路由器进行选择。（b）我们仅在单个冻结的 CLIP 中集成两组 soft prompt，并使用无需训练的路由器进行选择：一组侧重已见类别性能，另一组侧重未见类别泛化。
    }
    \label{figure:compare}
\end{figure}

\revision{
Our key innovation is a carefully designed \emph{dual-prompt architecture} for better balancing seen-class performance and unseen-class generalization. The architecture learns two complementary prompts using different data and training objectives, enabling them to focus on seen-class discrimination and unseen-class generalization, respectively; a training-free router then leverages their complementary strengths at inference.
}
\cn{
我们的核心创新是一种面向已见类性能与未见类泛化平衡而精心设计的双提示架构。该架构使用不同数据和训练目标学习两组互补提示，使其分别侧重已见类判别与未见类泛化；推理时，无需训练的路由器进一步发挥二者的互补优势。
}

More recently, several approaches~\cite{MoCoOp, MoPD, BeyondSoleStrength} explore prompt/model ensembling to improve generalization. However, they typically require multiple CLIP models or many prompt sets (Fig.~\ref{figure:compare} (a)), which increases computational cost, and often suffer from router training instability and load imbalance, and the resulting routing decisions are typically hard to interpret.
\cn{
近期也有一些方法 [MoCoOp, MoPD, BeyondSoleStrength]探索通过提示/模型集成提升泛化能力。然而，这类方法通常需要多个 CLIP 模型或大量提示集合（图~\ref{figure:compare}（a）），带来较高的计算开销；容易出现路由训练不稳定与负载不均衡等问题，且路由决策通常难以解释。
}

\begin{figure}[t]
    \centering
    \includegraphics[width=\columnwidth]{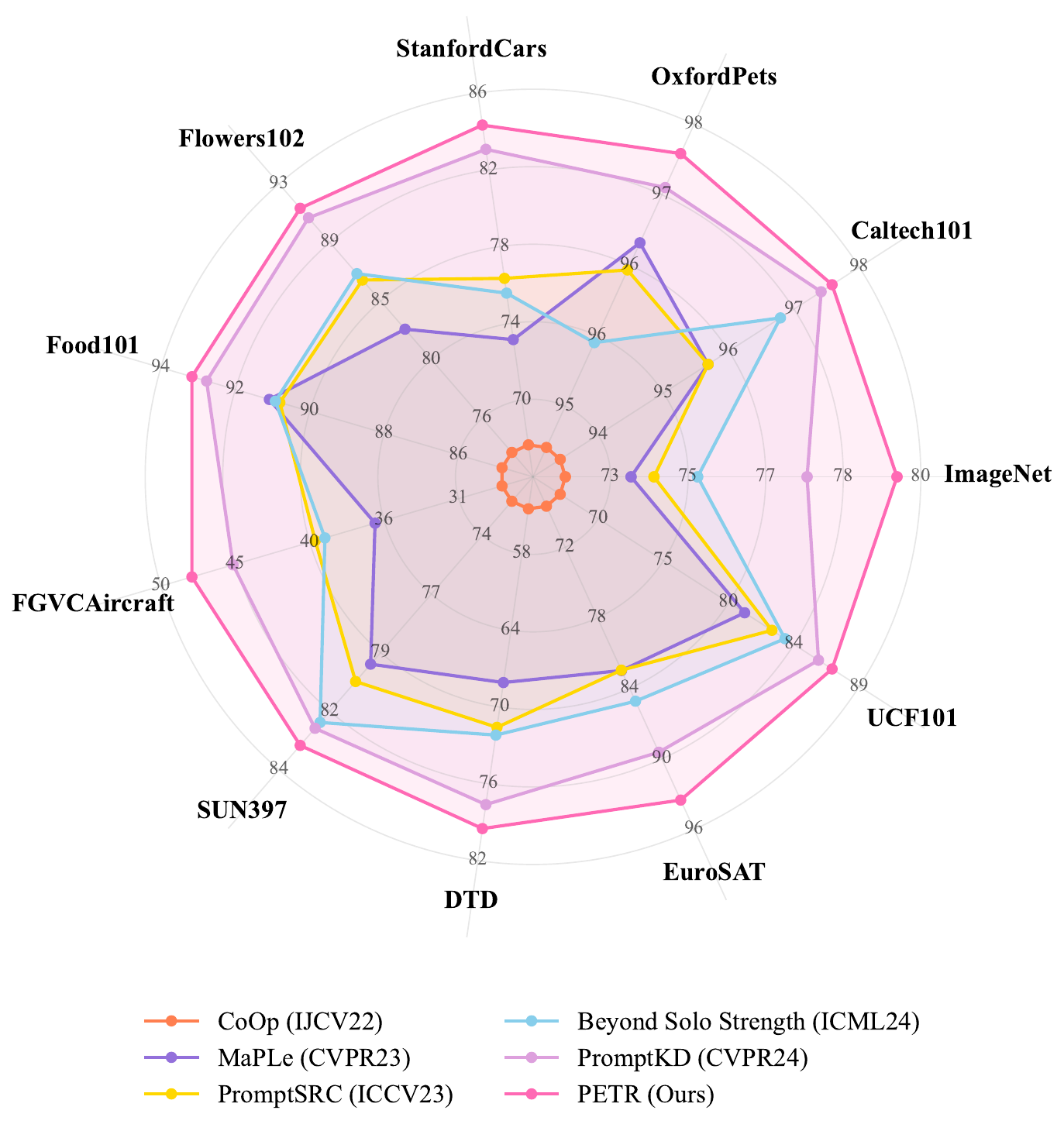}
    \caption{Harmonic mean (HM) performance comparison on base-to-novel generalization. PETR achieves state-of-the-art performance on 11 datasets.}
    \label{figure:performance comparison}
\end{figure}

\revision{
To address these challenges, we propose \textbf{PETR}, a general prompt ensembling framework that (i) ensembles only two prompt sets within a single frozen CLIP model and (ii) performs \emph{training-free, statistics-based} routing.
}
The routing is driven by an explicit distance-to-training-distribution score, providing an interpretable, likelihood-inspired signal and improved robustness under distribution shift (Fig.~\ref{figure:compare} (b)).
Extensive experiments on 11 benchmark datasets (Fig.~\ref{figure:performance comparison}) demonstrate consistent improvements on both seen and unseen classes.
\cn{
为了解决上述挑战，我们提出 PETR：一种通用的提示集成框架，（i）仅在单个冻结的 CLIP 模型内集成两组提示，（ii）并使用基于统计量的无需训练路由进行选择。该路由由显式的“与训练分布的距离”分数驱动，提供可解释、近似似然的路由信号，并在分布偏移下更稳健（图~\ref{figure:compare}（b））。我们在 11 个基准数据集上的实验（图~\ref{figure:performance comparison}）表明，PETR 在已见与未见类别上均能稳定提升。
}

The main contributions of this paper are as follows:
\begin{itemize}
\revision{
    \item We propose a prompt ensembling framework that explicitly constructs two role-specialized prompts for seen-class discrimination and unseen-class generalization within a single frozen CLIP model.
}
    \item The proposed training-free router with an \emph{interpretable} distance-to-training-distribution score, which avoids common MoE-style routing pathologies (router training instability and load imbalance) and can be plugged into existing prompt methods.
    \item We conduct extensive evaluations on 11 datasets, achieving new state-of-the-art results, demonstrating the effectiveness of the framework.
\end{itemize}
\cn{
本文的主要贡献如下：
\begin{itemize}
    \item 我们提出一个提示集成框架，在单个冻结的 CLIP 模型中显式构造两组职责专门化的提示，分别面向已见类判别与未见类泛化。
    \item 我们设计了一个无需训练的路由器，由可解释的“与训练分布的距离”分数驱动，避免路由训练不稳定与负载均衡问题，并可即插即用增强现有提示方法。
    \item 我们在 11 个数据集上进行了广泛评估，取得了新的最先进结果，证明了框架的有效性。
\end{itemize}
}

\section{Related Work}

\myparagraph{Prompt Learning in Vision-Language Models.}
Prompt learning \cite{prompt-tuning-1, prompt-tuning-2, prompt-tuning-3, tan2024compound} is a lightweight fine-tuning method that does not require large-scale training of the base model. By inserting a small number of learnable vectors before the model input, it offers advantages such as low parameter count and fast convergence.

CoOp~\cite{CoOp} fine-tunes CLIP~\cite{CLIP} by adding learnable prompts to the text branch, significantly improving performance on seen classes.
MaPLe~\cite{MaPLe} further improves prompt learning by adding learnable prompts to multiple Transformer layers in vision and language branches, enhancing both performance and generalization.
KgCoOp~\cite{KgCoOp} leverages hand-crafted prompts as prior knowledge to regularize learnable prompts.

\revision{
Unlike DCA~\cite{DCA}, which learns multiple peer prompts for diverse class representations, PETR uses different data and losses to construct two prompts specialized for seen-class discrimination and unseen-class generalization, and selects between them with a training-free router.
}
\cn{视觉-语言模型中的提示学习
提示学习 [prompt-tuning-1, prompt-tuning-2, prompt-tuning-3, tan2024compound] 是一种轻量级微调方法，无需对基础模型进行大规模训练。其做法是在模型输入前插入少量可学习向量，因此具备参数量小、收敛快等优点。

CoOp [CoOp] 通过在文本分支加入可学习提示对 CLIP [CLIP] 进行微调，显著提升已见类别的性能。
MaPLe [MaPLe] 进一步在多层 Transformer 中引入可学习提示，同时增强性能与泛化能力。
KgCoOp [KgCoOp] 将手工设计的提示作为先验知识，用于正则化可学习提示。

与 DCA [DCA] 学习多组地位对等的提示以获得多样化类别表示不同，PETR 使用不同的数据和损失构造两组分别专注于已见类判别与未见类泛化的提示，并通过无需训练的路由器在二者之间进行选择。
}

\myparagraph{Ensemble Learning.}
Ensemble learning \cite{ensemble-1, ensemble-2, ensemble-3} is a machine learning paradigm that combines the outputs of multiple models to achieve a better balance between performance and generalization. Some studies \cite{MoCoOp, MoPD, BeyondSoleStrength} attempt to combine ensemble learning with prompt learning.

However, existing prompt ensemble methods often require multiple CLIP models or many prompt sets, leading to substantial computational overhead. Moreover, their trainable routers/gates introduce additional optimization complexity and often suffer from MoE-style routing pathologies (router training instability and load imbalance), while the resulting routing decisions are typically hard to interpret.

\revision{
PETR differs by learning two prompts with distinct yet complementary roles from different data and objectives within a single frozen CLIP model, and selecting between them with a training-free, distribution-aware router.
}
\cn{集成学习
集成学习 [ensemble-1, ensemble-2, ensemble-3] 是一种通过组合多个模型输出来提升整体效果的机器学习范式，通常能够在性能与泛化之间取得更好的平衡。一些研究 [MoCoOp, MoPD, BeyondSoleStrength] 尝试将集成学习与提示学习结合。

然而，现有的提示集成方法往往需要多个 CLIP 模型或大量提示集合，计算开销较高；同时其可训练的路由/门控会引入额外的优化复杂度，常伴随路由训练不稳定与负载不均衡等 MoE 式问题，且路由决策通常难以解释。

PETR 的不同之处在于，它在单个冻结的 CLIP 模型中，利用不同数据和训练目标学习两组职责明确且互补的提示，并通过无需训练的分布感知路由器进行选择。
}

\myparagraph{Knowledge Distillation.}
Knowledge distillation \cite{distillation-1} is a method for training lightweight student models under the supervision of large pre-trained teacher models. There are many knowledge distillation studies on CLIP recently \cite{distillation-2,distillation-3}.

PromptKD \cite{PromptKD} performs prompt learning on a larger teacher network and distills its knowledge into the student prompts.
Benefiting from the advancements of the PromptKD model, we further explore using PromptKD's teacher network to distill two different student prompts using different data, and then combine their outputs.
\cn{知识蒸馏
知识蒸馏 [distillation-1] 是一种在大型预训练教师模型监督下训练轻量级学生模型的方法。
PromptKD [PromptKD] 在更大的教师网络上进行提示学习，并将其知识蒸馏到学生提示中。近年来，也有多项工作提出了对 PromptKD 的改进 [promptkd-1, promptkd-2]。
受 PromptKD 进展启发，我们进一步探索：使用 PromptKD 的教师网络，分别在不同数据上蒸馏两组不同的学生提示，并在推理阶段将它们的输出进行组合。
}

\begin{figure*}[t]
    \centering
    \includegraphics[width=\linewidth]{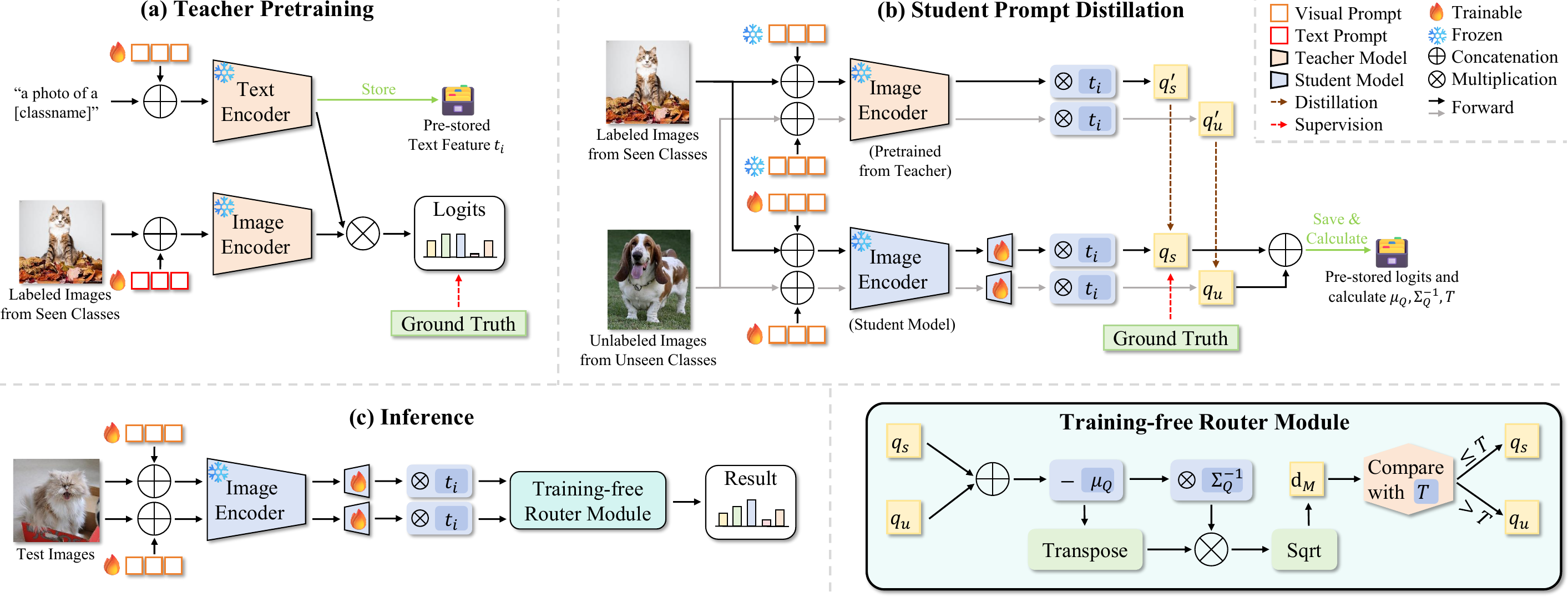}
    \caption{
    An overview of our proposed framework.
    (a) In the pre-training stage, we first train a teacher model and save all output text features $t_i$ for use in subsequent stages.
\revision{
    (b) In the distillation stage, we use labeled images from base classes and unlabeled images from novel classes, along with their respective loss functions, to distill the knowledge of the teacher model into two student prompts.
}
    Next, for all samples from seen classes, we store the concatenated vectors of the logits output by the two student models and compute $\mu_Q, \Sigma_Q, T$ for use in later stages.
    (c) In the inference stage, each sample is concatenated with the two student prompts and passed through the CLIP model to obtain two logits.
    These two logits are then processed by the Training-free Router Module to produce the final prediction.
    }
    \cn{
    框架概览。（a）预训练阶段：训练教师模型，并保存文本编码器输出的所有文本特征 t_i 供后续使用。（b）蒸馏阶段：分别使用已见类别的有标注图像与未见类别的无标注图像（及其对应损失），将教师模型知识蒸馏到两组学生提示中；随后对所有已见类别样本，存储两组学生输出 logits 的拼接向量，并计算 μ_Q、Σ_Q、T 供后续路由使用。（c）推理阶段：每个样本分别与两组学生提示结合，通过 CLIP 得到两组 logits，并由无需训练的路由模块生成最终预测。
}
    \label{figure:architecture}
\end{figure*}

\section{Method}

In this section, we begin by introducing the necessary background, then describe the main components of our proposed PETR framework, including teacher model pretraining, student prompt distillation, and inference process. Finally, we demonstrate how PETR can be integrated with existing methods.
\cn{
本节首先给出必要的背景知识，随后描述我们提出的 PETR 框架的关键组件，包括教师模型预训练、学生提示蒸馏与推理流程。最后，我们展示 PETR 如何与现有方法集成。
}

\subsection{Preliminaries}

\myparagraph{Vision-Language Models.}
The CLIP model \cite{CLIP} consists of an image encoder and a text encoder, which encode image and text inputs into high-dimensional feature vectors, respectively. CoOp \cite{CoOp} is a prompt learning method for CLIP models, which replaces the class template with $M$ learnable vectors. Prompt Ensembling uses $N$ prompt templates combined with $K$ class names and input into the text encoder to generate multiple text features $\left\{\textbf{t}_{ij}\mid i=1,\cdots,K;j=1,\cdots,N\right\}$. These text features are then averaged and compared with the image feature $\textbf{f}$ to get logits $\textbf{q}_i$:
\revision{
\begin{equation}
    \textbf{q}_i = \frac{\exp(\text{cos}(\textbf{f}, \bar{\textbf{t}}_i)/\gamma)}{\sum_{k=1}^K \exp(\text{cos}(\textbf{f}, \bar{\textbf{t}}_k)/\gamma)},
\label{equation:Prompt Ensembling}
\end{equation}
}
where $\bar{\textbf{t}}_i=\frac1N \sum_{j=1}^N \textbf{t}_{ij}$ is the average text feature for each class, $\text{cos}(\cdot,\cdot)$ is the cosine similarity function, and $\gamma$ is a learnable temperature parameter.
\cn{视觉-语言模型
CLIP [CLIP] 由图像编码器和文本编码器组成，分别将图像与文本输入编码为高维特征向量。
CoOp [CoOp] 通过将类别模板替换为 M 个可学习向量来实现提示学习。
Prompt Ensembling 使用 N 个提示模板与 K 个类别名组合输入文本编码器，生成多组文本特征 {t_ij | i=1~K; j=1,…,N}，并对每个类别做平均后与图像特征 f 进行对比得到 logits q_i：
q_i = exp(cos(f, t̄_i)/γ) / Σ_{k=1~K} exp(cos(f, t̄_k)/γ)
其中 t̄_i = (1/N) Σ_{j=1~N} t_{ij}；cos(·,·) 为余弦相似度；γ 为可学习的温度参数。
}

\myparagraph{PromptKD.}
PromptKD \cite{PromptKD} improves prompt-based fine-tuning performance using knowledge distillation.
It first trains a teacher model and keeps the textual embeddings.
Then, it distills a smaller image encoder from the teacher model through an unlabeled domain dataset.
The training objective is formulated as follows:
\begin{equation}
    \mathcal{L} = \alpha\, \text{KL}(\textbf{f}_t \textbf{W}^{\mathsf T},\textbf{f}_s \textbf{W}^{\mathsf T},\tau),
\label{equation:PromptKD}
\end{equation}
$\textbf{W}$ denotes the text features for each class output by the teacher model's text encoder, and $\textbf{f}_t$ and $\textbf{f}_s$ represent the image features from the teacher and student models, respectively.
$\text{KL}(\cdot, \cdot, \tau)$ is the knowledge distillation objective~\cite{distillation-1} with tempreture $\tau$.
$\alpha$ is the distillation weight that rescales the KD loss; its value may vary across datasets in PromptKD settings.
\cn{PromptKD
PromptKD [...] 通过知识蒸馏提升基于 prompt 的微调性能。
它首先训练一个教师模型并保留文本嵌入。
随后在无标注的域内数据集上，从教师模型蒸馏出一个更小的图像编码器（学生模型）。
训练目标可写为：

L = α KL(f_t W^{\mathsf T}, f_s W^{\mathsf T}, τ)

这里，W 表示教师模型文本编码器为每个类别输出的文本特征；f_t 和 f_s 分别表示教师与学生模型的图像特征。
KL(·,·,τ) 是温度系数 τ 的知识蒸馏目标 [distillation-1]。
这里 α 为蒸馏权重，到对 KD 损失进行缩放的作用；在 PromptKD 的设置中，其取值在不同数据集上可能不同。
}

\subsection{PETR}
\label{sec:petr}
The overview of our framework is shown in Fig.~\ref{figure:architecture}.
\revision{
PETR addresses the long-standing difficulty of simultaneously maintaining strong seen-class performance and unseen-class generalization in prompt learning. Its key innovation is a carefully designed dual-prompt architecture, where two complementary prompts focus on these two requirements separately and a training-free router selects between them at inference.
This design improves unseen-class generalization while preserving strong seen-class discrimination, leading to a better overall balance.
}
We will provide a detailed explanation in the following part.

\myparagraph{Teacher Pretraining.}
During the teacher model pre-training stage illustrated in Fig.~\ref{figure:architecture} (a), we train a prompt on teacher model. For all classes, we save the corresponding text features $t_i$ output from the text encoder for use in subsequent stages.
\cn{
我们框架的概述如图~\ref{figure:architecture} 所示。
PETR 旨在解决提示学习中长期存在的难题，即如何同时保持较强的已见类性能与未见类泛化能力。其核心创新是精心设计的双提示架构：两组互补提示分别侧重这两方面的能力，并在推理时通过无需训练的路由器进行选择。
该设计在保持已见类判别能力的同时改善了未见类泛化，从而获得更加均衡的整体性能。
下文将对各个模块进行详细说明。

教师模型预训练
在教师模型预训练阶段，我们训练一个教师模型上的提示（图~\ref{figure:architecture}（a））。对于所有类别，保存文本编码器输出的对应文本特征 t_i，供后续阶段使用。
}

\myparagraph{Student Prompt Distillation.}
During the distillation phase illustrated in Fig.~\ref{figure:architecture} (b), we train two sets of prompt learning parameters based on a frozen CLIP model.
The first prompt $\theta_s$ is specifically designed to perform well on seen classes, while the second prompt $\theta_u$ is tailored for better generalization to unseen classes (subscripts $s/u$ denote \emph{seen/unseen}, respectively).
For each sample $x$, the corresponding logits outputs $\textbf{q}_s, \textbf{q}_u$ and their concatenation $\tilde{\textbf{q}}$ are defined as follows.
\begin{equation}
\begin{aligned}
    	\textbf{q}_s &= \operatorname{CLIP}(x,\theta_s),\\
    	\textbf{q}_u &= \operatorname{CLIP}(x,\theta_u), \\
    	\tilde{\textbf{q}} &= \left[\textbf{q}_s, \textbf{q}_u\right].
\end{aligned}
\label{equation:logits}
\end{equation}
\cn{
在图~\ref{figure:architecture}（b）所示的蒸馏阶段，我们基一个冻结的 CLIP 模型训练两组提示参数。
其中，第一组提示 θ_s 旨在提升已见类别性能；第二组提示 θ_u 则更强调对未见类别的泛化（下标 s/u 分别表示 seen/unseen）。
对于每个样本 x，对应 logits q_s, q_u 和拼接结果 q̃ 定义为：
q_s = CLIP(x, θ_s)
q_u = CLIP(x, θ_u)
q̃ = [q_s, q_u]
}

Define $Q$ as the set of concatenated logits $\tilde{\mathbf{q}}$ collected from training images of seen classes. We then compute the feature vectors for all \emph{seen-class} training samples from $Q$ to obtain the mean $\mu_Q$, covariance matrix $\Sigma_Q$, and threshold $T$, which are used for subsequent calculations.
\revision{
\begin{equation}
\begin{aligned}
   \mu_Q&=\frac 1 {\lvert Q \rvert} \sum_{\tilde{\textbf{q}}\in Q} \tilde{\textbf{q}},\\
   \Sigma_Q&=\frac 1 {\lvert Q \rvert} \sum_{\tilde{\textbf{q}}\in Q} \left(\tilde{\textbf{q}}-\mu_Q\right) \left(\tilde{\textbf{q}}-\mu_Q\right)^{\mathsf T},\\
    T &= \lambda \cdot \mathop{\operatorname{top1\%}}\limits_{\tilde{\mathbf{q}} \in Q}\left\{ d_M\left(\tilde{\textbf{q}},Q\right) \right\},\\
    \text{where } d_M(\tilde{\mathbf{q}},Q) &= \sqrt{(\tilde{\mathbf{q}} - \mu_Q)^{\mathsf T} \Sigma_Q^{-1} (\tilde{\mathbf{q}} - \mu_Q)},
\label{equation:mean and covariance}
\end{aligned}
\end{equation}
$d_M(\tilde{\mathbf{q}},Q)$ is the Mahalanobis distance between the concatenated logit vector $\tilde{\mathbf{q}}$ and the seen-class distribution $Q$.
$\operatorname{top1\%}$ denotes the 99th percentile of the Mahalanobis distances over $Q$, i.e., the cutoff above which the largest 1\% of distances lie.
$\lambda$ is a threshold scaling hyperparameter shared across all datasets that controls the routing boundary between seen and unseen classes.

The Mahalanobis distance provides a likelihood-based measure of deviation from the seen-class training distribution and is a classic OOD score for deep features~\cite{Lee2018MahalanobisOOD}.
Under a multivariate Gaussian approximation, $d_M^2(\tilde{\mathbf{q}},Q)$ corresponds to the sample-dependent term of the negative log-likelihood up to a constant~\cite{mahalanobis-1,Lee2018MahalanobisOOD}.
Thus, a larger distance indicates a lower likelihood under the seen-class distribution and provides a principled OOD routing signal.
Unlike Euclidean distance, the Mahalanobis distance accounts for dimension-wise scales and cross-dimensional correlations through $\Sigma_Q$.
Estimating these statistics in the low-dimensional logit space further improves the stability of covariance estimation and inversion in practice.
}
\cn{
定义 Q 为从已见类别训练图像收集到的拼接 logits 集合 q̃。我们然后对集合 Q 中所有已见类别训练样本的特征向量进行统计，得到均值 μ_Q、协方差矩阵 Σ_Q 和阈值 T，用于后续计算。
μ_Q = (1/|Q|) · ∑_{q̃∈Q} q̃
Σ_Q = (1/|Q|) · ∑_{q̃∈Q} (q̃-μ_Q)(q̃-μ_Q)^{\mathsf T}
T = λ · top1
其中 d_M(q̃, Q) 表示拼接 logit 向量 q̃ 相对于已见类别分布 Q 的马氏距离。
top1\% 表示 Q 中马氏距离的第 99 百分位数，即仅有最大的 1\% 距离高于该截断值。
λ 是所有数据集共用的阈值缩放超参数，用于控制已见类与未见类之间的路由判别边界。

马氏距离能够从似然角度衡量样本偏离已见类训练分布的程度，也是深度特征上经典的 OOD 评分 [Lee2018MahalanobisOOD]。
在多元高斯近似下，d_M²(q̃, Q) 对应负对数似然中与样本相关的项，仅相差一个常数 [mahalanobis-1, Lee2018MahalanobisOOD]。
因此，距离越大，表示样本在已见类分布下的似然越低，从而为 OOD 路由提供具有统计依据的判别信号。
与欧氏距离不同，马氏距离通过 Σ_Q 同时校正不同维度的尺度及维度间相关性。
此外，在低维 logit 空间中估计这些统计量，也使协方差估计与求逆在实践中更加稳定。
}

\myparagraph{Inference.}
During the inference phase illustrated in Fig.~\ref{figure:architecture}(c), each sample is concatenated with both $\theta_s$ and $\theta_u$ and passed through the student CLIP model to obtain two logits. These two logits are then processed by the training-free router module to produce the final prediction.

The training-free router module determines whether the sample belongs to a seen or unseen class based on $\tilde{\textbf{q}}$.
\revision{
If the sample is classified as a seen class, the final prediction is the class with the largest logit in $\textbf{q}_s$; otherwise, it is the class with the largest logit in $\textbf{q}_u$.
}
\cn{
在图~\ref{figure:architecture}（c）所示的推理阶段，每个样本分别与 θ_s 和 θ_u 结合，并通过学生 CLIP 模型得到两组 logits。随后，这两组 logits 由无需训练的路由模块处理，得到最终预测。

无需训练的路由模块根据 q̃ 判断样本更像已见类别还是未见类别。若判为已见类别，则将 q_s 中 logit 最大的类别作为最终预测；否则将 q_u 中 logit 最大的类别作为最终预测。
}

\revision{
The router dynamically selects the final prediction $\mathcal{R}$ according to the comparison between the Mahalanobis distance of the concatenated logits and the threshold $T$.
}
All quantities used by the Mahalanobis score, including $\mu_Q, \Sigma_Q$, and the threshold $T$, are precomputed once from seen-class training samples and stored.
This computation of router is carried out in the low-dimensional logit space and is typically negligible compared to the CLIP forward pass.
The process is formulated as follows:
\revision{
\begin{equation}
\mathcal{R} =
\begin{cases}
    \underset{c}{\arg\max}\,[\textbf{q}_s]_c, & d_M(\tilde{\textbf{q}}, Q) \leq T \\
    \underset{c}{\arg\max}\,[\textbf{q}_u]_c, & \text{otherwise}
\end{cases}
\label{equation:PETR}
\end{equation}
where $c$ indexes the candidate classes and $[\textbf{q}]_c$ denotes the logit associated with class $c$.
}
\cn{
路由器根据拼接 logits 的马氏距离与阈值 T 的比较结果，动态确定最终预测 R。
马氏距离评分所需的量，包括 μ_Q、Σ_Q 以及阈值 T，会在训练阶段基于已见类别样本一次性统计并存储。
路由器的计算发生在低维 logits 空间中，相比 CLIP 的前向传播开销通常可以忽略。
过程如下所述：
R =
{ argmax_c [q_s]_c, 若 d_M(q̃, Q) ≤ T
  argmax_c [q_u]_c, 否则 }
其中 c 为候选类别索引，[q]_c 表示类别 c 对应的 logit。
}

\subsection{Plug-and-Play Integration}

\myparagraph{PETR + PromptKD}.
We can combine PETR with PromptKD. During the teacher model pretraining stage, we train a teacher model based on CLIP ViT-L/14 using the same training setup as PromptKD. In the student prompt distillation stage, we distill two student prompts with the same training setup as PromptKD, both sharing a single CLIP ViT-B/16 model.

The student prompt $\theta_u$ is distilled using unlabeled images from unseen classes to improve generalization to unseen categories, while the student prompt $\theta_s$ is distilled using labeled images from seen classes to enhance classification performance on seen categories.
\revision{
The two prompts are intentionally optimized with different data and objectives to induce their role specialization.
}
The loss function is defined as follows:
\begin{equation}
\begin{aligned}
    \mathcal{L}_\text{u} &= \alpha \text{KL}(\textbf{f}_t \textbf{W}^{\mathsf T},\textbf{f}_s \textbf{W}^{\mathsf T},\tau), \\
    \mathcal{L}_\text{s} &= \alpha \text{KL}(\textbf{f}_t \textbf{W}^{\mathsf T},\textbf{f}_s \textbf{W}^{\mathsf T},\tau) + \text{CE}(y, \textbf{f}_s \textbf{W}^{\mathsf T}),
\label{equation:PETR + PromptKD}
\end{aligned}
\end{equation}
where $y$ is image label, $\alpha$ weights the KD term relative to the CE term when distilling the seen-class prompt $\theta_s$. For the unseen prompt $\theta_u$, we use KD only; thus $\alpha$ mainly rescales the KD gradients.
\cn{PETR + PromptKD
我们将 PETR 与 PromptKD 结合使用。教师模型预训练阶段，我们采用与 PromptKD 相同的训练设置，在 CLIP ViT-L/14 上训练教师模型。

学生提示蒸馏阶段，我们同样沿用 PromptKD 的训练设置，在共享的 CLIP ViT-B/16 上蒸馏两组学生提示：$\theta_u$ 使用来自未见类别的无标注图像以提升泛化；$\theta_s$ 使用来自已见类别的有标注图像以增强已见类别分类性能。两组提示分别使用不同的数据与训练目标进行优化，从而形成职责专门化。损失定义为：
$L_u = α KL(f_t W^{\mathsf T}, f_s W^{\mathsf T}, τ)$
$L_s = α KL(f_t W^{\mathsf T}, f_s W^{\mathsf T}, τ) + CE(y, f_s W^{\mathsf T})$
其中 y 是真实标签，α 与 PromptKD 中的含义一致，用于缩放 KD 目标。在蒸馏已见提示 θ_s 时，α 进一步控制 KD 相对 CE 的强度；在蒸馏未见提示 θ_u（仅 KD）时，α 主要用于缩放 KD 梯度。
}

\myparagraph{PETR+CoOp.}  
Our PETR framework can be directly integrated with the CoOp method. Specifically, we skip the teacher model pretraining stage and directly use the soft prompt trained by CoOp as $\theta_s$, and the original CLIP hard prompt as $\theta_u$. Both prompts share the same CLIP model, differing only in the text prompts. Unless otherwise specified, PETR in the following refers to PETR + PromptKD.
\cn{
PETR+CoOp
PETR 也可以直接与 CoOp 结合。具体来说，我们跳过教师模型预训练阶段，直接使用 CoOp 训练得到的 soft prompt 作为 $\theta_s$，并使用原始 CLIP 的 hard prompt 作为 $\theta_u$。
两组提示共享同一个 CLIP 模型，仅文本提示不同。除非另有说明，下文中的 PETR 默认指 PETR + PromptKD。
}


\section{Results}

We conduct extensive evaluations of the proposed methods and compare them with existing approaches. The following presents the main experimental settings and result analysis.
\cn{
我们对提出的方法进行了广泛的评估，并与现有方法进行了比较。以下是主要的实验设置和结果分析。
}

\subsection{Settings}

\myparagraph{Training Data Usage.}
We summarize the dataset usage into four stages (a--d).
Stages (a) and (b) describe the PromptKD setting (included for comparison), while stages (c) and (d) correspond to PETR.

(a) Teacher pretraining (PromptKD): the teacher is trained on labeled seen-class training data, i.e., \textbf{seen images} with \textbf{seen labels}.
(b) Student distillation (PromptKD, for comparison): the student is distilled using all unlabeled training images.
In this setting, because all seen images have already been used in (a), stage (b) implicitly reveals which images belong to unseen classes;
equivalently, the unlabeled images in (b) can be viewed as a mixture of \textbf{seen images} and \textbf{unseen images}.

(c) PETR distillation: we reuse the PromptKD teacher and distill two students using different subsets of the training data.
One student uses \textbf{seen images} with \textbf{seen labels}, and the other uses unlabeled \textbf{unseen images}.
(d) Router statistics estimation: we use unlabeled \textbf{seen images} only, feed them into both students, concatenate their outputs, and compute the logits statistics required by our training-free router.

\begin{figure}[t]
    \centering
    \includegraphics[width=\columnwidth]{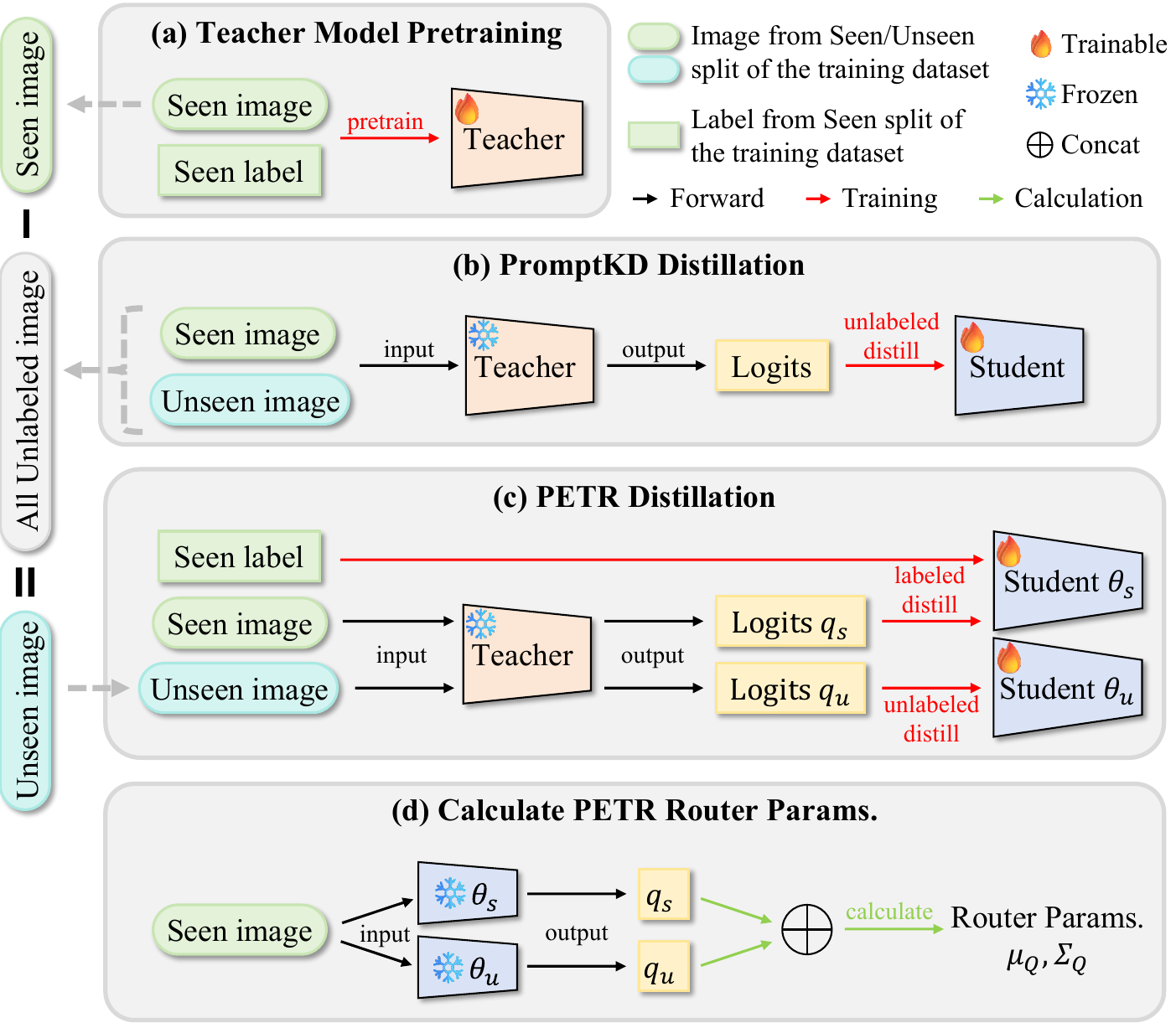}
    \caption{Dataset usage in our experimental setting in four stages (a--d).
    (a) PromptKD Teacher pretraining: labeled seen images with seen labels.
    (b) PromptKD Student distillation for comparison: all unlabeled training images, which can be viewed as seen images + unseen images.
    (c) PETR distillation: one student uses seen images with seen labels, and the other uses unlabeled unseen images.
    (d) Router statistics estimation: unlabeled seen images only.
    Overall, the dataset usage in PETR is consistent with PromptKD.}
    \cn{实验设置中的数据使用（a--d 四阶段）
    (a) 教师预训练：seen images + seen labels。
    (b) 学生蒸馏（PromptKD，用于对比）：训练集全部无标注图像，可视作 seen images + unseen images。
    (c) PETR 蒸馏：一个学生使用带标注的 seen images（seen images + seen labels），另一个学生使用无标注的 unseen images。
    (d) 路由统计量估计：仅使用无标注的 seen images。
    因此，本文实验设置与 PromptKD 的数据使用设置一致。
    }
    \label{figure:dataset_usage}
\end{figure}

Figure~\ref{figure:dataset_usage} visualizes stages (a--d) and the corresponding data used in each stage. Notably, PETR may appear to use a subset of \textbf{unseen images} in stage (c) that is not explicitly separated in the PromptKD description. However, under the PromptKD data protocol, this subset is already implicitly determined: it is the set difference between (i) the unlabeled images used for distillation in stage (b) and (ii) the seen-class images used for supervised teacher pretraining in stage (a). Formally,
\[
\{\text{unseen images}\} = \{\text{all unlabeled images}\} - \{\text{seen images}\}.
\]

Existing prompt learning methods such as CoOp and MaPLe typically follow the \emph{inductive zero-shot learning} setting: they are trained only on labeled images from seen classes, without accessing unseen-class images or any unlabeled target-distribution data during training. In contrast, PromptKD follows the \emph{transductive zero-shot learning} setting: while never accessing unseen labels, it can use unlabeled training images that include unseen-class samples during distillation.

Many prior works in transductive zero-shot learning \cite{use-unseen-images-1, use-unseen-images-2, use-unseen-images-3} commonly assume that it is known whether an unlabeled image belongs to the seen-class pool or the unseen-class pool, and design subset-specific learning strategies accordingly. Therefore, our practice of explicitly distinguishing and utilizing seen vs.\ unseen images during distillation also falls within the transductive zero-shot learning paradigm.

\revision{
Therefore, PETR and PromptKD use the same training data and supervision signals. In practical category-expansion scenarios, unlabeled images collected for a predefined set of new classes naturally constitute the unseen-class pool, so this partition requires no additional annotation.
}
\cn{ 训练数据使用
我们将数据使用过程划分为 a--d 四个阶段。
其中 (a)(b) 为 PromptKD 的设置（用于对比），(c)(d) 为本文 PETR 的设置。

(a) 教师预训练（PromptKD）：使用已见（Base）类别训练集中的带标注图像训练教师模型，即 seen images + seen labels。
(b) 学生蒸馏（PromptKD，用于对比）：使用训练集中的所有无标注图像进行蒸馏。
在该设置下，由于阶段 (a) 已使用过全部 seen images，因此阶段 (b) 在事实上隐含了“哪些图像属于 unseen 类别”的信息；
等价地，(b) 中的无标注图像可视作由 seen images 与 unseen images 组成。

(c) PETR 蒸馏：复用 PromptKD 的教师模型，在训练集的不同子集上蒸馏两组学生。
其中一个蒸馏使用带标注的已见图像（seen images + seen labels），另一个蒸馏使用无标注的未见图像（unseen images）。
(d) 路由统计量估计：仅使用无标注的 seen images，将其输入两个学生模型并拼接输出，计算路由所需的 logits 统计量（如均值与协方差）。

图~\ref{figure:dataset_usage} 对 a--d 四阶段及其所用数据做了可视化总结。需要说明的是，表面上看 PETR 在阶段 (c) “单独使用了”在 PromptKD 描述中未显式区分的 unseen images；但在 PromptKD 的数据使用协议下，该子集在数据划分层面实际上已经被隐式确定：阶段 (b) 使用的全体无标注图像与阶段 (a) 中已用于监督预训练的已见类别图像之间做差，即可得到未见类别图像集合。形式化地，有
\text{unseen images} \triangleq \text{all unlabeled images} \mathbin{\backslash} \text{seen images}.

在现有的提示学习方法中，代表性的 CoOp、MaPLe 等通常工作在 inductive zero-shot learning 设定下：模型只在来自已见类别的带标注图像上进行训练，训练过程中既不访问来自未见类别的图像，也不利用任何测试阶段的无标注数据，因此无法从目标分布中获取额外信息。与之相对，PromptKD 采用的是 transductive zero-shot learning 设定：在始终不访问未见类别标签的前提下，它在蒸馏阶段可以使用包含未见类别样本在内的无标注训练图像。

在许多以往的 transductive zero-shot 学习工作 [...] 中，研究者往往默认知道哪些无标注样本来自已见类别、哪些来自未见类别，并在此基础上对不同子集设计专门的学习策略。因此，本文在蒸馏阶段显式区分并利用已见与未见图像的做法，与既有的 transductive zero-shot learning 范式是一致且合理的。

因此，PETR 与 PromptKD 使用相同的训练数据与监督信息；在实际的类别扩展场景中，面向预先定义的新类别所收集的无标签图像自然构成未见类样本池，因而该划分不需要任何额外标注。
}

\myparagraph{Base-to-novel Generalization.}
Following previous studies \cite{CoOp, CoCoOp}, we split the training and test datasets into base and novel classes. We pretrain the teacher model following the same training setup as PromptKD \cite{PromptKD}. We distill two student prompts using the labeled training set of base classes and the unlabeled training sets of the novel classes, respectively.
\cn{
基础到新类别泛化
我们遵循以往工作 [CoOp, CoCoOp]，将训练与测试类别划分为基础类（base）与新类（novel）。教师模型按照 PromptKD 的训练设置进行预训练 [PromptKD]。
随后，我们分别使用 base 类别的有标注训练集与 novel 类别的无标注训练集，蒸馏两组学生提示。
}

\myparagraph{Cross-dataset Evaluation.}
We follow the cross-dataset evaluation training setup of PromptKD \cite{PromptKD}. The teacher model is pretrained on the ImageNet \cite{ImageNet} dataset using a 16-shot training data configuration. Then, the student prompt is trained on the unlabeled training set of the target dataset and evaluated on the test set to assess cross-dataset performance.
\cn{跨数据集评估
我们遵循 PromptKD 的跨数据集评估训练设置 [PromptKD]。教师模型在 ImageNet [ImageNet] 上以 16-shot 配置进行预训练。
随后，学生提示在目标数据集的无标注训练集上训练，并在测试集上评估跨数据集性能。
}

\myparagraph{Domain Generalization.}
We also follow the domain generalization training setup of PromptKD. The teacher model is pretrained on the ImageNet dataset, and then the student prompt is trained on the unlabeled training set of the target dataset and evaluated on the test set to assess domain generalization performance.
\cn{领域泛化
我们同样遵循 PromptKD 的领域泛化训练设置 [PromptKD]。教师模型在 ImageNet 上进行预训练。
随后，学生提示在目标数据集的无标注训练集上训练，并在测试集上评估领域泛化性能。
}

\myparagraph{PETR+CoOp Performance.}
We further applied the PETR framework to CoOp and conducted main experiments: base-to-novel generalization, cross-dataset evaluation, and domain generalization.
For the base-to-novel experiment, we follow the training setup of CoOp. For the other two experiments, we use the prompt ensembling approach. Prompts used for each dataset are provided in the Supplemental Materials.
\cn{
PETR+CoOp 性能
我们进一步将 PETR 应用于 CoOp，并进行了主要实验：基础到新类别泛化、跨数据集评估与领域泛化。
其中，基础到新类别泛化遵循 CoOp 的训练设置；另外两种设置使用 prompt ensembling 的评估方式。各数据集使用的具体 prompts 见附录。
}

\myparagraph{Datasets.}
Following CoOp and CoCoOp \cite{CoCoOp}, we use 11 image recognition datasets to evaluate base-to-novel generalization and cross-dataset evaluation: general object classification datasets ImageNet \cite{ImageNet} and Caltech101 \cite{Caltech101}, fine-grained classification datasets OxfordPets \cite{OxfordPets}, StanfordCars \cite{StanfordCars}, Flowers102 \cite{OxfordFlowers}, Food101 \cite{Food101}, and FGVCAircraft \cite{FGVCAircraft}, scene recognition dataset SUN397 \cite{SUN397}, action recognition dataset UCF101 \cite{UCF101}, texture classification dataset DTD \cite{DTD}, and satellite image classification dataset EuroSAT \cite{EuroSAT}. For cross-dataset evaluation, ImageNet is used as the source dataset, and out-of-distribution performance is tested on the target datasets ImageNet-A \cite{ImageNet-A}, ImageNet-R \cite{ImageNet-R}, ImageNet-Sketch \cite{ImageNet-Sketch}, and ImageNetV2 \cite{ImageNet-V2}.
\cn{
数据集
我们遵循 CoOp 与 CoCoOp [CoOp, CoCoOp]，选择了11个广泛使用的图像识别数据集评估基础到新颖的泛化和跨数据集评估：通用对象分类数据集ImageNet [14]和Caltech101[15]，细粒度分类数据集OxfordPets  [16]、StanfordCars [17]、Flowers102 [18]、Food101 [19]和 FGVCAircraft [20]，场景识别数据集SUN397 [21]，动作识别数据集UCF101 [22]，纹理分类数据集DTD [23]，卫星图像分类数据集EuroSAT [24]。跨数据集评估使用ImageNet 作为源数据集，并在目标数据集ImageNet-A [25]、ImageNet-R [26]、ImageNet-Sketch [27]和 ImageNetV2 [28]上测试分布外性能。
}

\myparagraph{Training Details.}
We report the accuracy on base and novel classes, as well as the harmonic mean (HM) of these two accuracies. Results of base and novel class accuracy are averaged over 3 runs.

For PETR+CoOp, we use the ViT-B/16 CLIP model. The context length is fixed at 4, and the context vectors are initialized with the word embeddings of the pre-trained phrase ``a photo of a''. The batch size is 32, and the number of epochs is 200. All experiments are conducted on a single V100 GPU.

For PETR + PromptKD, we use the ViT-L/14 CLIP model as the teacher and the ViT-B/16 CLIP model as the target student. The prompt depth for PETR + PromptKD is 9, the visual and language prompt lengths are 4, and the SGD optimizer is used. To ensure a fair comparison, the number of samples used for student model distillation is consistent with PromptKD, with 40 training epochs, a batch size of 8, and a learning rate of 0.005. We follow the same data augmentation scheme as PromptKD. Prompt vectors are randomly initialized from a normal distribution, and the first layer text prompt is initialized with the word embeddings of ``a photo of a''. All experiments are conducted on a V100 GPU.
\cn{
训练细节
我们报告 base 与 novel 类别的准确率，以及两者的调和平均（HM）。base 与 novel 的结果均为 3 次运行的平均值。

在PETR+CoOp设置中使用ViT-B/16 的 CLIP 模型。我们将上下文长度固定为 4，并使用预训练的短语“a photo of a”的词嵌入对上下文向量进行初始化。批量大小为 32，共 200 个周期。所有实验均在一张 V100 GPU上运行。

在PETR + PromptKD设置中使用 ViT-L/14 CLIP 模型作为教师模型，使用 ViT-B/16 CLIP 模型作为目标学生模型。PETR + PromptKD的提示深度为 9，视觉和语言提示长度为 4，使用SGD优化器。为保证公平比较，所有学生模型蒸馏所需的样本量与PromptKD保持一致，均训练 40 个周期，批量大小为 8，学习率为 0.005。我们遵循与 PromptKD 相同的数据增强方案。提示向量随机初始化为正态分布，第一层的文本提示使用 "a photo of a" 的词嵌入初始化。所有实验均在一张 V100 GPU上运行。
}

\subsection{Domain Generalization}
Table~\ref{table:Domain generalization} evaluates the direct transferability of models trained on ImageNet to various out-of-domain datasets. On the target datasets, PETR outperforms previous methods on 3 out of 4 datasets, and also achieves a higher average accuracy. The inferior performance on the ImageNet V2 dataset may be due to its subtle differences from the original ImageNet distribution, which are difficult to capture.
\cn{
表~\ref{table:Domain generalization} 评估了在 ImageNet 上训练的模型向域外数据集迁移的能力。
在目标数据集上，PETR 在 4 个数据集中的 3 个上优于以往方法，并取得更高的平均准确率。
在 ImageNet-V2 上略低的表现可能源于其与原始 ImageNet 分布差异较为细微，难以被统计量充分捕捉。
}

\begin{table}[ht]
    \caption{Domain generalization results. PETR outperforms previous methods on 3 datasets, and achieves higher average accuracy.}
    \cn{
    领域泛化结果。PETR 在 3 个数据集上优于以往方法，并取得更高的平均准确率。
    }
    \small \centering
    \setlength{\tabcolsep}{5.6pt}
    \begin{tabular}{lccccc}
        \hline\noalign{\smallskip}
        \textbf{Method} & \textbf{-V2} & \textbf{-Sketch} & \textbf{-A} & \textbf{-R} & \textbf{Avg.} \\
        \hline\noalign{\smallskip}
        CoOp & 64.20 & 47.99  & 49.71  & 75.21  & 59.28 \\
        CoCoOp  & 64.07 & 48.75 & 50.63 & 76.18 & 59.91 \\
        MaPLe & 64.07 & 49.15 & 50.90 & 76.98 & 60.27 \\
        MoPD-MMP & 65.40 & 49.30 & 50.70 & 77.10 & 60.63 \\
        PromptSRC & 64.35 & 49.55 & 50.90 & 77.80 & 60.65 \\
        MoCoOp & \revision{\textbf{75.88}} & 48.97 & 46.50 & 61.31 & 58.17 \\
        BeyondSoleStrength & 65.73 & 50.70 & 52.11 & 78.11 & 61.66 \\
        PromptKD & \revision{69.77} & 58.72 & 70.36 & \revision{87.01} & 71.47 \\
        \hline\noalign{\smallskip}
        \rowcolor{tabhighlight} PETR (ours) & \revision{70.39} & \revision{\textbf{60.10}} & \revision{\textbf{71.52}} & \revision{\textbf{87.77}} & \revision{\textbf{72.44}} \\
    \end{tabular}
    \label{table:Domain generalization}
\end{table}

\begin{table*}[!t]
\caption{
Performance of our proposed PETR method on multiple datasets, compared with existing approaches.
\revision{
The table reports results for base classes, novel classes, and HM.
}
$\Delta$ denotes the performance improvement of our methods over the compared methods.
\revision{
PETR achieves superior performance on all 11 datasets, with an overall HM improvement of 1.41\% over PromptKD.
}
}
\cn{
表 1. 我们提出的 PETR 在多个数据集上的性能，并与现有方法进行对比。
表中报告基础类、新类及其 HM。Δ 表示相对对比方法的提升幅度。
PETR 在 11 个数据集上均取得更优结果，整体 HM 相对 PromptKD 提升 1.41\%。
}
\small \centering
    \setlength{\tabcolsep}{4pt}
    \begin{tabular}{lcc|ccc|ccc|ccc}
    \toprule
    & \textbf{Ensemble} & \multirow{2}{*}{\textbf{Transductive}} & \multicolumn{3}{c|}{\textbf{Average}} & \multicolumn{3}{c|}{\textbf{ImageNet}} & \multicolumn{3}{c}{\textbf{Caltech101}}  \\
    & \textbf{Method} & & \textbf{Base} & \textbf{Novel} & \textbf{HM} & \textbf{Base} & \textbf{Novel} & \textbf{HM} & \textbf{Base} & \textbf{Novel} & \textbf{HM} \\
    \midrule
    CoOp~\cite{CoOp} & & & 82.69  & 63.22  & 71.66  & 76.47  & 67.88  & 71.92  & 98.00  & 89.81  & 93.73 \\
    CoCoOp~\cite{CoCoOp} & & & 80.47  & 71.69  & 75.83  & 75.98  & 70.43  & 73.10  & 97.96  & 93.81  & 95.84  \\
    MaPLe~\cite{MaPLe} & & & 82.28 & 75.14 & 78.55  & 76.66 & 70.54 & 73.47 & 97.74 & 94.36 & 96.02 \\
    MoPD-MMP~\cite{MoPD} & \cmark & & 83.43 & 75.41 & 79.22 & 77.60 & 69.91 & 73.55 & 98.36 & 94.18 & 96.22 \\
    PromptSRC~\cite{PromptSRC} & & & 84.26 & 76.10 & 79.97 & 77.60 & 70.73 & 74.01 & 98.10 & 94.03 & 96.02 \\
    MoCoOp~\cite{MoCoOp} & \cmark & & 83.32 & 77.34 & 80.17 & 76.52 & 69.20 & 72.67 & 98.43 & 94.87 & 96.61 \\
    BeyondSoleStrength~\cite{BeyondSoleStrength} & \cmark & & 85.48 & 77.17 & 81.11 & 78.74 & 71.68 & 75.04 & 98.58 & 95.74 & 97.14 \\
    PromptKD~\cite{PromptKD} & & \cmark & 86.96 & 80.73 & 83.73 & 80.83 & 74.66 & 77.62 & 98.91 & 96.65 & 97.77 \\
    \midrule
    \rowcolor{tabhighlight} PETR (ours) & \cmark & \cmark & \revision{\textbf{88.42}} & \revision{\textbf{82.09}} & \revision{\textbf{85.14}} & \revision{\textbf{84.36}} & \revision{\textbf{75.60}} & \revision{\textbf{79.74}} & \textbf{99.18} & \textbf{96.98} & \textbf{98.07} \\
    $\Delta$ & & & \incre{\revision{+1.46}} & \incre{\revision{+1.36}} & \incre{\revision{+1.41}} & \incre{\revision{+3.53}} & \incre{\revision{+0.94}} & \incre{\revision{+2.12}} & \incre{+0.27} & \incre{+0.33} & \incre{+0.30} \\
    \bottomrule
    \end{tabular}
    \vspace{0.7em}\\
    \begin{tabular}{lcc|ccc|ccc|ccc}
    \toprule
    & \textbf{Ensemble} & \multirow{2}{*}{\textbf{Transductive}} & \multicolumn{3}{c|}{\textbf{OxfordPets}} & \multicolumn{3}{c|}{\textbf{StanfordCars}} & \multicolumn{3}{c}{\textbf{Flowers102}}  \\
    & \textbf{Method} & & \textbf{Base} & \textbf{Novel} & \textbf{HM} & \textbf{Base} & \textbf{Novel} & \textbf{HM} & \textbf{Base} & \textbf{Novel} & \textbf{HM} \\
    \midrule
    CoOp~\cite{CoOp} & & & 93.67 & 95.29 & 94.47 & 78.12 & 60.40 & 68.13 & 97.60 & 59.67 & 74.06 \\
    CoCoOp~\cite{CoCoOp} & & & 95.20 & 97.69 & 96.43 & 70.49 & 73.59 & 72.01 & 94.87 & 71.75 & 81.71 \\
    MaPLe~\cite{MaPLe} & & & 95.43 & 97.76 & 96.58 & 72.94 & 74.00 & 73.47 & 95.92 & 72.46 & 82.56 \\
    MoPD-MMP~\cite{MoPD} & \cmark & & 95.43 & 97.78 & 96.59 & 79.75 & 73.06 & 76.26 & 97.28 & 75.18 & 84.81 \\
    PromptSRC~\cite{PromptSRC} & & & 95.33 & 97.30 & 96.30 & 78.27 & 74.97 & 76.58 & 98.07 & 76.50 & 85.95 \\
    MoCoOp~\cite{MoCoOp} & \cmark & & 95.59 & 96.64 & 96.11 & 76.34 & 73.26 & 74.77 & 97.18 & 77.21 & 86.05 \\
    BeyondSoleStrength~\cite{BeyondSoleStrength} & \cmark & & 95.96 & 97.71 & 96.83 & 81.26 & 78.48 & 79.85 & 98.77 & 77.52 & 86.86 \\
    PromptKD~\cite{PromptKD} & & \cmark & 96.30 & 98.01 & 97.15 & 82.80 & 83.37 & 83.13 & 99.42 & 82.62 & 90.24 \\
    \midrule
    \rowcolor{tabhighlight} PETR (ours) & \cmark & \cmark & \textbf{96.79} & \textbf{98.23} & \textbf{97.50} & \textbf{84.22} & \textbf{84.51} & \textbf{84.36} & \textbf{99.46} & \textbf{83.69} & \textbf{90.90} \\
    $\Delta$ & & & \incre{+0.49} & \incre{+0.22} & \incre{+0.35} & \incre{+1.42} & \incre{+1.14} & \incre{+1.23} & \incre{+0.04} & \incre{+1.07} & \incre{+0.66} \\
    \bottomrule
    \end{tabular}
    \vspace{0.7em}\\
    \begin{tabular}{lcc|ccc|ccc|ccc}
    \toprule
    & \textbf{Ensemble} & \multirow{2}{*}{\textbf{Transductive}} & \multicolumn{3}{c|}{\textbf{Food101}} & \multicolumn{3}{c|}{\textbf{FGVCAircraft}} & \multicolumn{3}{c}{\textbf{SUN397}}  \\
    & \textbf{Method} & & \textbf{Base} & \textbf{Novel} & \textbf{HM} & \textbf{Base} & \textbf{Novel} & \textbf{HM} & \textbf{Base} & \textbf{Novel} & \textbf{HM} \\
    \midrule
    CoOp~\cite{CoOp} & & & 88.33 & 82.26 & 85.19 & 40.44 & 22.30 & 28.75 & 80.60 & 65.89 & 72.51 \\
    CoCoOp~\cite{CoCoOp} & & & 90.70 & 91.29 & 90.99 & 33.41 & 23.71 & 27.74 & 79.74 & 76.86 & 78.27 \\
    MaPLe~\cite{MaPLe} & & & 90.71 & 92.05 & 91.38 & 37.44 & 35.61 & 36.50 & 80.82 & 78.70 & 79.75 \\
    MoPD-MMP~\cite{MoPD} & \cmark & & 90.85 & 91.75 & 91.30 & 41.18 & 35.55 & 38.16 & 82.20 & 78.54 & 80.33 \\
    PromptSRC~\cite{PromptSRC} & & & 90.67 & 91.53 & 91.10 & 42.73 & 37.87 & 40.15 & 82.67 & 78.47 & 80.52 \\
    MoCoOp~\cite{MoCoOp} & \cmark & & 90.25 & 91.57 & 90.90 & 38.78 & 38.09 & 38.43 & 81.43 & 77.45 & 79.39 \\
    BeyondSoleStrength~\cite{BeyondSoleStrength} & \cmark & & 90.75 & 91.70 & 91.22 & 43.22 & 36.47 & 39.56 & 83.88 & 80.86 & 82.34 \\
    PromptKD~\cite{PromptKD} & & \cmark & 92.43 & 93.68 & 93.05 & 49.12 & 41.81 & 45.17 & 83.69 & 81.54 & 82.60 \\
    \midrule
    \rowcolor{tabhighlight} PETR (ours) & \cmark & \cmark & \textbf{93.04} & \textbf{93.96} & \textbf{93.50} & \textbf{54.10} & \textbf{42.63} & \textbf{47.68} & \textbf{84.88} & \textbf{81.89} & \textbf{83.36} \\
    $\Delta$ & & & \incre{+0.61} & \incre{+0.28} & \incre{+0.45} & \incre{+4.98} & \incre{+0.82} & \incre{+2.51} & \incre{+1.19} & \incre{+0.35} & \incre{+0.76} \\
    \bottomrule
    \end{tabular}
    \vspace{0.7em}\\
    \begin{tabular}{lcc|ccc|ccc|ccc}
    \toprule
    & \textbf{Ensemble} & \multirow{2}{*}{\textbf{Transductive}} & \multicolumn{3}{c|}{\textbf{DTD}} & \multicolumn{3}{c|}{\textbf{EuroSAT}} & \multicolumn{3}{c}{\textbf{UCF101}}  \\
    & \textbf{Method}  & & \textbf{Base} & \textbf{Novel} & \textbf{HM} & \textbf{Base} & \textbf{Novel} & \textbf{HM} & \textbf{Base} & \textbf{Novel} & \textbf{HM} \\
    \midrule
    CoOp~\cite{CoOp} & & & 79.44 & 41.18 & 54.24 & 92.19 & 54.74 & 68.69 & 84.69 & 56.05 & 67.46 \\
    CoCoOp~\cite{CoCoOp} & & & 77.01 & 56.00 & 64.85 & 87.49 & 60.04 & 71.21 & 82.33 & 73.45 & 77.64 \\
    MaPLe~\cite{MaPLe} & & & 80.36 & 59.18 & 68.16 & 94.07 & 73.23 & 82.35 & 83.00 & 78.66 & 80.77 \\
    MoPD-MMP~\cite{MoPD} & \cmark & & 81.40 & 57.65 & 67.50 & 86.79 & 78.20 & 82.27 & 86.85 & 77.68 & 82.01 \\
    PromptSRC~\cite{PromptSRC} & & & 83.37 & 62.97 & 71.75 & 92.90 & 73.90 & 82.32 & 87.10 & 78.80 & 82.74 \\
    MoCoOp~\cite{MoCoOp} & \cmark & & 81.94 & 60.99 & 69.93 & 94.79 & 85.18 & 89.73 & 85.28 & 79.31 & 82.17 \\
    BeyondSoleStrength~\cite{BeyondSoleStrength} & \cmark & & 84.72 & 63.16 & 72.37 & 95.52 & 76.41 & 84.90 & 88.83 & 79.10 & 83.68 \\
    PromptKD~\cite{PromptKD} & & \cmark & 85.84 & 71.37 & 77.94 & 97.54 & 82.08 & 89.14 & 89.71 & 82.27 & 86.10 \\
    \midrule
    \rowcolor{tabhighlight} PETR (ours) & \cmark & \cmark & \textbf{88.39} & \textbf{72.83} & \textbf{79.86} & \textbf{97.55} & \textbf{89.07} & \textbf{93.12} & \textbf{90.68} & \textbf{83.76} & \textbf{87.08} \\
    $\Delta$ & & & \incre{+2.55} & \incre{+1.46} & \incre{+1.92} & \incre{+0.01} & \incre{+6.99} & \incre{+3.15} & \incre{+0.97} & \incre{+1.49} & \incre{+0.98} \\
    \bottomrule
    \end{tabular}
\label{table:Base2Novel}
\end{table*}

\subsection{Base-to-novel Generalization}
\label{sec:base2new}
As shown in Table~\ref{table:Base2Novel}, we compare the proposed PETR with state-of-the-art prompt learning methods across 11 recognition datasets.
\revision{
Compared to previous works, PETR achieves superior performance on all 11 datasets, with an overall HM improvement of 1.41\% over PromptKD.
}
For ImageNet, FGVCAircraft, and DTD, the base class performance improves substantially because our method distills a prompt using labeled base class images.
\revision{
For EuroSAT, the novel class performance is greatly improved. This is because the distribution of novel classes in EuroSAT is distinct from the base classes, prompt $\theta_u$ can generalize better to novel categories.
}
\cn{
如表~\ref{table:Base2Novel} 所示，我们在 11 个识别数据集上将 PETR 与当前先进的提示学习方法进行比较。
相较以往方法，PETR 在全部 11 个数据集上均取得更优结果，整体 HM 相对 PromptKD 提升 1.41\%。

在 ImageNet、FGVCAircraft 与 DTD 上，Base 类别性能提升尤为明显，主要来自我们利用有标注的 Base 图像蒸馏提示。
在 EuroSAT 上，Novel 类别性能提升显著，这是因为 EuroSAT 的 Novel 类别分布与 Base 差异较大，提示 $\theta_u$ 更易泛化到新类别。
}

\begin{table*}[t]
    \caption{
    Cross-dataset benchmark evaluation.
\revision{
    PETR outperforms previous methods on 9 out of 10 datasets and attains the highest average performance.
    MoCoOp~\cite{MoCoOp} is not included because its original paper does not report cross-dataset evaluation results.
}
    }
    \cn{
    跨数据集评估结果。PETR 在 10 个数据集中的 9 个上优于以往方法，并取得最高的平均性能。由于 MoCoOp 原论文未报告跨数据集评估结果，因此未将其纳入比较。
    }
    \small \centering
    \setlength{\tabcolsep}{4pt}
    \begin{tabular}{lcc|ccccccccccc}
        \hline\noalign{\smallskip}
        Method & \makecell{\textbf{Ensemble}\\\textbf{Method}} & \textbf{Transductive} & \rotatebox{90}{\textbf{Caltech101}} & \rotatebox{90}{\textbf{OxfordPets}} & \rotatebox{90}{\textbf{StanfordCars}} & \rotatebox{90}{\textbf{Flowers102}} & \rotatebox{90}{\textbf{Food101}} & \rotatebox{90}{\textbf{FGVCAircraft}} & \rotatebox{90}{\textbf{SUN397}} & \rotatebox{90}{\textbf{DTD}} & \rotatebox{90}{\textbf{EuroSAT}} & \rotatebox{90}{\textbf{UCF101}} & \rotatebox{90}{\textbf{Avg.}} \\
        \hline\noalign{\smallskip}
        CoOp~\cite{CoOp}       & & & 93.70 & 89.14 & 64.51 & 68.71 & 85.30 & 18.47 & 64.15 & 41.92 & 46.39 & 66.55 & 63.88 \\
        CoCoOp~\cite{CoCoOp} & & & \textbf{94.43} & 90.14 & 65.32 & 71.88 & 86.06 & 22.94 & 67.36 & 45.73 & 45.37 & 68.21 & 65.74 \\
        MaPLe~\cite{MaPLe} & & & 93.53 & 90.49 & 65.57 & 72.23 & 86.20 & 24.74 & 67.01 & 46.49 & 48.06 & 68.69 & 66.30 \\
        MoPD-MMP~\cite{MoPD} & \cmark & & 94.30 & 90.70 & 65.70 & 71.10 & 86.30 & 23.40 & 66.90 & 46.40 & 48.30 & 68.90 & 66.20 \\
        PromptSRC~\cite{PromptSRC} & & & 93.60 & 90.25 & 65.70 & 70.25 & 86.15 & 23.90 & 67.10 & 46.87 & 45.50 & 68.75 & 65.81 \\
        BeyondSoleStrength~\cite{BeyondSoleStrength} & \cmark & & 93.91 & 90.72 & 71.94 & 72.59 & 86.68 & 26.03 & 66.07 & 49.31 & 48.18 & 69.14 & 67.46 \\
        PromptKD~\cite{PromptKD} & & \cmark & 93.61 & 91.59 & 73.93 & 75.33 & 88.84 & 26.24 & 68.57 & 55.08 & 63.74 & 76.39 & 71.33 \\
        \hline\noalign{\smallskip}
        \rowcolor{tabhighlight} PETR (ours) & \cmark & \cmark & \revision{94.28} & \revision{\textbf{93.32}} & \revision{\textbf{75.48}} & \revision{\textbf{79.78}} & \revision{\textbf{89.45}} & \revision{\textbf{29.34}} & \revision{\textbf{69.25}} & \revision{\textbf{58.33}} & \revision{\textbf{65.04}} & \revision{\textbf{76.66}} & \revision{\textbf{73.09}} \\
    \end{tabular}
    \label{table:Cross-dataset evaluation} 
\end{table*}

\subsection{Cross-dataset Evaluation}
In Table~\ref{table:Cross-dataset evaluation}, we compare the cross-dataset performance of PETR with previous methods.
\revision{
Compared with previous approaches, our method outperforms others on 9 out of 10 datasets and attains the highest average performance.
On Caltech101, PETR remains comparable to the best-performing method.
The slightly lower performance may be attributed to the high visual similarity between certain classes (e.g., Leopards vs. Cougar face), which poses additional challenges for cross-dataset evaluation.
}
\cn{
在表~\ref{table:Cross-dataset evaluation} 中，我们比较了 PETR 与以往方法的跨数据集性能。
相较以往方法，PETR 在 10 个数据集中的 9 个上表现更好，并取得最高的平均性能。

PETR 在 Caltech101 上的结果仍与最佳方法接近。
其略低的结果可能源于部分类别间高度视觉相似（例如 Leopards vs. Cougar face），使跨数据集评估更具挑战。
}

\revision{
\section{Discussion}
}

\revision{
\subsection{Extension to CoOp}
}
Due to space limitations, only the average performance is reported in this section. Detailed experimental results for PETR+CoOp are provided in the Supplemental Materials. Our approach consistently outperformed CoOp significantly in Table~\ref{table:PETR+CoOp}. These results demonstrate the generality of the PETR framework, showing that it can enhance the performance of prompt learning methods.

In the base-to-novel generalization experiments, PETR+CoOp achieves slightly lower performance than CoOp on base classes. This is due to the classification performance loss introduced by the router.
\revision{
However, PETR+CoOp shows a substantial improvement on Novel and HM performace, benefiting from the superior generalization ability of the original CLIP prompt compared to CoOp.
}
In the cross-dataset evaluation and domain generalization experiments, PETR+CoOp also achieves better results.
\cn{
由于篇幅限制，本节仅报告 PETR+CoOp 在三个设置下的平均性能；更完整的结果见补充材料。
如表~\ref{table:PETR+CoOp} 所示，PETR+CoOp 相比 CoOp 均取得了显著提升，验证了 PETR 的通用性，能够增强现有提示学习方法。

在基础到新类别泛化设置下，PETR+CoOp 在 Base 上略低于 CoOp，主要来自路由器引入的少量分类损失。
但在 Novel 与 HM 上提升明显，这得益于原始 CLIP prompt 相比 CoOp 具有更强的泛化能力。
在跨数据集评估与领域泛化设置下，PETR+CoOp 也取得了更优异的成绩。
}

\begin{table}[ht]
    \caption{Performance of PETR+CoOp on multiple benchmarks. PETR+CoOp achieves significantly better base-to-novel performance compared to CoOp, and also outperforms CoOp on the other two benchmarks.}
    \cn{
    PETR+CoOp 在多个设置下的性能对比。
    Base-to-novel 部分报告 Base / Novel / HM 的平均结果；Cross-dataset 与 Domain Generalization 部分报告各自基准上多数据集的平均结果。
    PETR+CoOp 在 Base-to-novel 设置下显著优于 CoOp，并在另外两个设置下也取得小幅提升。
    }
    \small \centering
    \begin{tabular}{lc|ccc}
    \toprule
    & & \textbf{CoOp} & \textbf{PETR+CoOp} & {$\Delta$} \\
    \midrule
    \multirow{3}{*}{\shortstack[l]{\textbf{Base-to-}\\\textbf{novel}}} 
        & \textbf{Base} & \textbf{82.69} & 82.39 & \decre{-0.30} \\
        & \revision{\textbf{Novel}}  & 63.22 & \textbf{73.82} & \incre{+10.60} \\
        & \textbf{HM}   & 71.66 & \textbf{77.49} & \incre{+5.83} \\
    \midrule
    \multicolumn{2}{l|}{\textbf{Cross-dataset}} & \multirow{2}{*}{63.88} & \multirow{2}{*}{\textbf{65.70}} & \multirow{2}{*}{\incre{+1.82}} \\
    \multicolumn{2}{l|}{\textbf{Evaluation}} \\
    \midrule
    \multicolumn{2}{l|}{\textbf{Domain}} & \multirow{2}{*}{59.28} & \multirow{2}{*}{\textbf{59.63}} & \multirow{2}{*}{\incre{+0.35}} \\
    \multicolumn{2}{l|}{\textbf{Generalization}} \\
    \bottomrule
    \end{tabular}
    \label{table:PETR+CoOp}
\end{table}

\begin{table*}[ht]
    \caption{When distilling the student prompt $\theta_s$, we use both the KL loss with the teacher model outputs and the CE loss with ground-truth labels, resulting in improved performance on the base classes. We conducted extensive experiments on the base class data, and results show that incorporating CE loss with ground-truth labels can improve model performance.}
    \cn{
    蒸馏学生提示 $\theta_s$ 时，同时使用与教师输出对齐的 KL 损失和与真实标签对齐的 CE 损失，可提升 Base 类别性能。
    我们在 Base 数据上进行了广泛实验，结果表明引入 CE 损失能够进一步提升模型表现。
    }
    \small \centering
    \begin{tabular}{lcccccccccccc}
    \toprule
        \textbf{Method} & \rotatebox{90}{\textbf{ImageNet}} & \rotatebox{90}{\textbf{Caltech101}} & \rotatebox{90}{\textbf{OxfordPets}} & \rotatebox{90}{\textbf{StanfordCars}} & \rotatebox{90}{\textbf{Flowers102}} & \rotatebox{90}{\textbf{Food101}} & \rotatebox{90}{\textbf{FGVCAircraft}} & \rotatebox{90}{\textbf{SUN397}} & \rotatebox{90}{\textbf{DTD}} & \rotatebox{90}{\textbf{EuroSAT}} & \rotatebox{90}{\textbf{UCF101}} & \rotatebox{90}{\textbf{Avg.}} \\
    \midrule
        Only KL Loss& 81.86 & \textbf{99.23}& 96.49 & 83.47& \textbf{99.49} & 92.73 & 50.52 & 83.98 & 84.88 & 97.06 & 89.95 & 87.78 \\
        \rowcolor{tabhighlight} KL+CE Loss &  \textbf{\revision{84.36}}&  99.18 &  \textbf{96.79}&  \textbf{84.22}&  99.46&  \textbf{93.04}&  \textbf{54.10}&  \textbf{84.88}& \textbf{88.39}& \textbf{97.55}& \textbf{90.68} & \textbf{\revision{88.42}}\\
        $\Delta$& \incre{\revision{+2.50}} & \decre{-0.05} & \incre{+0.30} & \incre{+0.75} & \decre{-0.03} & \incre{+0.31} & \incre{+3.58} & \incre{+0.90} & \incre{+3.51} & \incre{+0.49} & \incre{+0.73} & \incre{\revision{+0.64}} \\
    \bottomrule
    \end{tabular}
    \label{table:Distillation Method}
\end{table*}

\revision{
\subsection{Distillation Method}
}
The PromptKD paper only uses the KD loss to distill the student prompt. In contrast, PETR’s prompt $\theta_s$ is distilled using images from seen classes, allowing the use of both the KD loss with teacher model outputs and the CE loss with ground-truth labels. As shown in Table~\ref{table:Distillation Method}, incorporating the CE loss leads to notable performance improvements, especially on datasets such as ImageNet, FGVCAircraft, and DTD, where the base class performance improvements are most significant in the base-to-novel generalization results. This highlights the crucial role of labeled data for prompt distillation, and supports our analysis in Section~\ref{sec:base2new}.
\cn{ 蒸馏方法
PromptKD 仅使用 KD 损失蒸馏学生提示。
而 PETR 的已见提示 θ_s 使用已见类别的有标注图像进行蒸馏，因此可以同时使用 KD 损失与带标签的 CE 损失。
表~\ref{table:Distillation Method} 展示了引入 CE 损失后的性能提升，尤其在 ImageNet、FGVCAircraft 与 DTD 等数据集上更为显著。
这些结果进一步说明：有标注数据对提示蒸馏至关重要，也与第~\ref{sec:base2new} 节的分析相呼应。
}

\begin{figure*}[t]
    \centering
    \includegraphics[width=\linewidth]{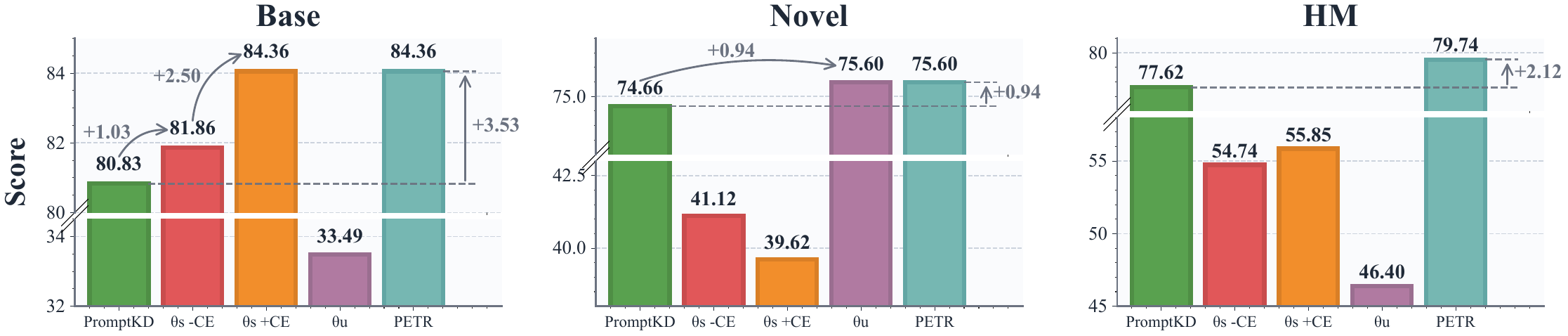}
    \caption{
\revision{
    Ablation on distillation data, CE supervision, and routing under the Base/Novel split on ImageNet dataset.
}
    Distilling $\theta_s$ using \textbf{seen images} improves \textbf{Base}, and adding label-supervised CE loss further strengthens $\theta_s$ on \textbf{Base}.
\revision{
    Distilling $\theta_u$ using \textbf{unseen images} improves \textbf{Novel} but hurts \textbf{Base}, revealing complementary specialization between $\theta_s$ and $\theta_u$.
}
    PETR's training-free router is used to get a more balanced harmonic mean.
    }
    \cn{
    在 ImageNet 数据集上 Base/Novel 划分下关于蒸馏数据、CE 监督与路由的消融结果。
    使用 \textbf{seen images} 蒸馏 $\theta_s$ 可提升 \textbf{Base}，并且加入基于标签的 CE 损失会进一步增强 $\theta_s$ 的 \textbf{Base} 性能。
    使用 \textbf{unseen images} 蒸馏 $\theta_u$ 可提升 \textbf{Novel}，但会损伤 \textbf{Base}，表明 $\theta_s$ 与 $\theta_u$ 具有互补专门化。
    PETR 的训练无关路由器用于获得更均衡的调和平均（HM）。
    }
    \label{fig:without_router}
\end{figure*}

\revision{
\subsection{Role of the Router}
}
Fig.~\ref{fig:without_router} provides a ablation on ImageNet dataset to disentangle the effects of (i) the prompt is distilled on seen or unseen images and (ii) whether to use CE loss when distilling the seen-class prompt.
Compared to PromptKD, which distills a single prompt on \emph{all} unlabeled images, PETR distills two prompts from the same PromptKD teacher using disjoint data:
$\theta_s$ is distilled on \emph{seen images} (with labels), while $\theta_u$ is distilled on \emph{unseen images} (without labels).

\revision{
This separation induces strong specialization: $\theta_s$ improves substantially on \textbf{Base} but drops sharply on \textbf{Novel}, whereas $\theta_u$ improves on \textbf{Novel} but degrades on \textbf{Base}.
}
Moreover, distilling $\theta_s$ with an additional CE loss further boosts \textbf{Base} performance by providing direct class-discriminative guidance beyond teacher matching.

To recover balanced performance, PETR employs the training-free router.
\revision{
At test time, the router predicts whether a sample is more likely \textbf{Base} (in-distribution) or \textbf{Novel} (OOD), and accordingly selects $\mathbf{q}_s$ or $\mathbf{q}_u$ as the final output.
}
\cn{路由器的作用
图~\ref{fig:without_router} 给出了一个在 ImageNet 数据集上的消融对比，用来拆解两点影响：（i）在 seen or unseen 上蒸馏 prompt，以及（ii）在蒸馏已见类别 prompt 时是否引入利用 CE 监督。
与 PromptKD 在\emph{所有}无标注图像上蒸馏单一 prompt 不同，PETR 从同一个 PromptKD 教师模型中，用互不重叠的数据蒸馏出两组 prompt：
$\theta_s$ 仅在\emph{已见图像}上蒸馏（可用标签），而 $\theta_u$ 仅在\emph{未见图像}上蒸馏（无标签）。

这种数据划分会带来明显的专门化：$\theta_s$ 在 \textbf{Base} 上显著提升，但在 \textbf{Novel} 上大幅下降；相对地，$\theta_u$ 在 \textbf{Novel} 上提升，但在 \textbf{Base} 上明显下降。
此外，我们在 $\theta_s$ 蒸馏中加入与真实标签对齐的 CE 损失，能够提供直接的类别判别信号，从而进一步提升 \textbf{Base} 性能。

为获得更均衡的整体表现，PETR 使用前文介绍的训练无关路由器：基于已见类别训练样本 logits 统计量计算的马氏距离 OOD 检测器。
测试时路由器判断样本更像 \textbf{Base}（域内）还是 \textbf{Novel}（OOD），并相应选择 $\mathbf{q}_s$ 或 $\mathbf{q}_u$ 作为最终输出。
}

\begin{figure*}[!t]
\revision{
    \centering
    \includegraphics[width=\linewidth]{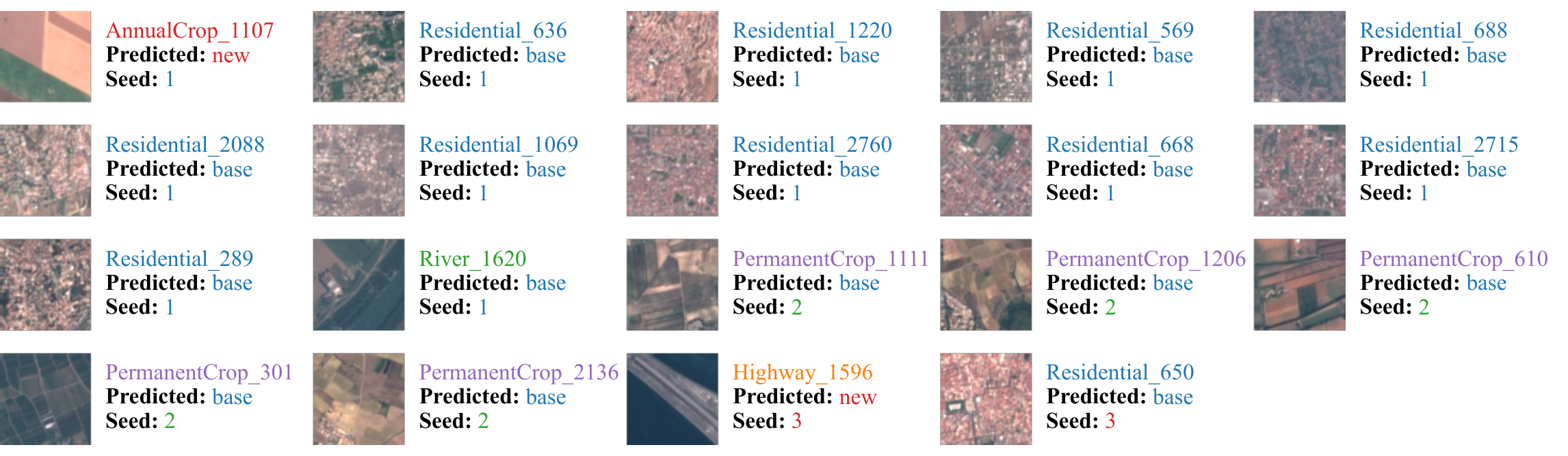}
    \caption{All 19 incorrectly routed EuroSAT samples collected across three random seeds.
    Each sample is annotated with its class name, predicted routing branch, and random seed.}
    \cn{
    三个随机种子下 EuroSAT 中全部 19 个错误路由样本。
    每个样本均标注了类别名称、预测路由分支和随机种子。
    }
    \label{fig:wrong_images}
}
\end{figure*}

\revision{
\subsection{Failure-Case Analysis}
\cn{失败案例分析}
}

\revision{
We further examine routing stability under the standard base-to-novel class split by collecting all incorrectly routed samples across three random seeds.
Incorrect routing occurs only on EuroSAT, where 19 errors are observed among 24300 test predictions aggregated over the three seeds, corresponding to a routing accuracy of 99.92\%.
Figure~\ref{fig:wrong_images} presents all these errors, while the remaining ten datasets achieve perfect routing under every seed.
}
\cn{
我们收集了标准 base-to-novel 类别划分下三个随机种子的全部错误路由样本，以进一步考察路由稳定性。
错误仅出现在 EuroSAT：汇总三个随机种子的测试结果后，共有 24300 次预测，其中 19 次路由错误，路由准确率为 99.92\%。
图~\ref{fig:wrong_images} 展示了全部错误样本，而其余十个数据集在每个随机种子下均未出现路由错误。
}

\revision{
Among the 19 errors, 17 Novel samples are routed to the Base branch, whereas only two Base samples are routed to the Novel branch.
The Novel-to-Base errors mainly involve Residential (11 samples) and PermanentCrop (5 samples), together with one River sample.
These satellite images contain repetitive textures and mixed local land-cover patterns, which may place their logit representations close to the routing boundary.
The two Base-to-Novel errors come from AnnualCrop and Highway.
Thus, the routing errors are confined to a few boundary cases, supporting the overall reliability and stability of the router.
}
\cn{
在 19 个错误样本中，17 个 Novel 样本被路由至 Base 分支，而仅有两个 Base 样本被路由至 Novel 分支。
Novel-to-Base 错误主要来自 Residential（11 个）和 PermanentCrop（5 个），另有一个来自 River。
这些卫星图像包含重复纹理和混合的局部地物结构，可能使其 logit 表征接近路由边界。
两个 Base-to-Novel 错误样本分别来自 AnnualCrop 和 Highway。
因此，路由错误仅局限于少量边界样本，进一步表明该路由器具有较高的总体可靠性与稳定性。
}

\begin{figure*}[!t]
\revision{
    \centering
    \includegraphics[width=\linewidth]{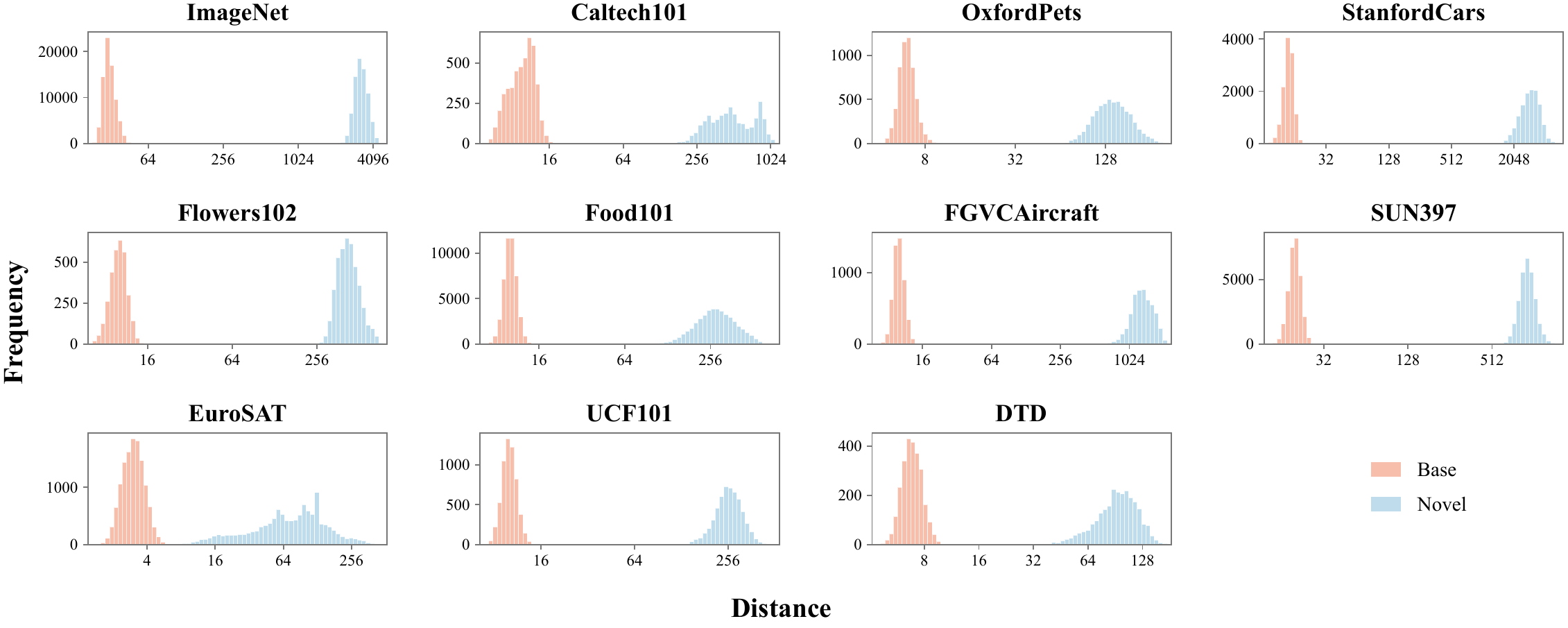}
    \caption{Mahalanobis-distance distributions of Base and Novel samples on all 11 datasets.
    The two distributions are clearly separated on ten datasets, while only EuroSAT exhibits a small tail overlap.
    For clearer visualization, all horizontal axes are shown on a log$_2$ scale.}
    \cn{
    全部 11 个数据集上 Base 与 Novel 样本的马氏距离分布。
    两个分布在其中 10 个数据集上间隔明显，仅在 EuroSAT 上存在少量尾部重叠。
    为便于观察，所有横坐标均采用 $\log_2$ 尺度。
    }
    \label{fig:mahalanobis_distributions}
    \par\addvspace{2\baselineskip}
    
    \includegraphics[width=\linewidth]{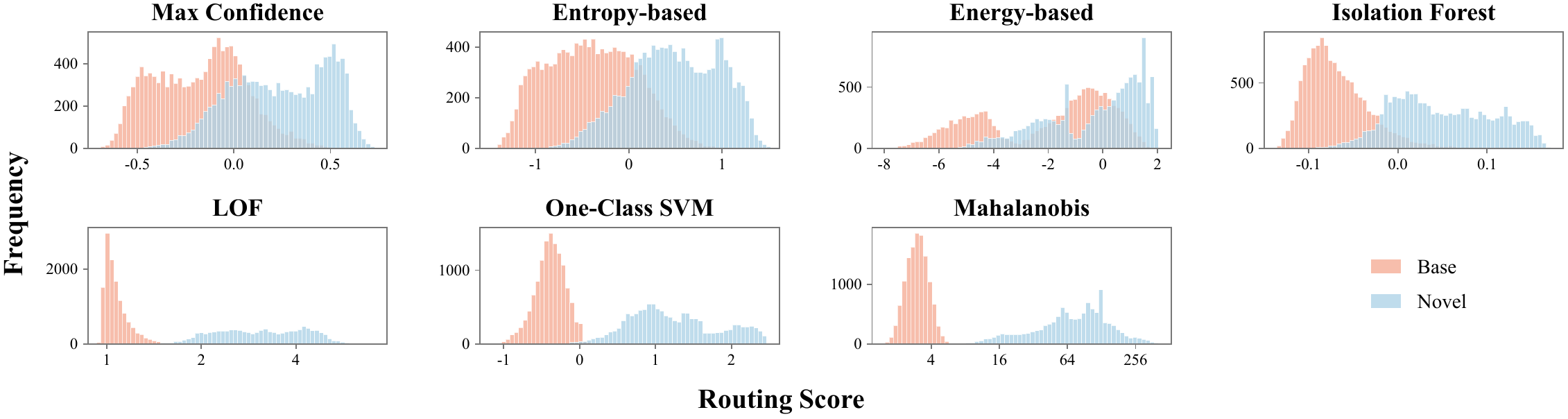}
    \caption{Routing-score distributions of seven routing methods on EuroSAT.
    LOF, One-Class SVM, and Mahalanobis provide clearer separation between Base and Novel samples than the other methods, with Mahalanobis exhibiting the clearest overall separation.
    For clearer visualization, the horizontal axes of LOF and Mahalanobis are shown on a log$_2$ scale, while the others use linear scales.}
    \cn{
    七种路由方法在 EuroSAT 上的分数分布。
    相比其他方法，LOF、One-Class SVM 和 Mahalanobis 能够更清晰地区分 Base 与 Novel 样本，其中 Mahalanobis 的总体分离效果最为明显。
    为便于观察，LOF 和 Mahalanobis 的横坐标采用 $\log_2$ 尺度，其余方法采用线性尺度。
    }
    \label{fig:eurosat_routing_distributions}
    \par\addvspace{2\baselineskip}
    
    \includegraphics[width=\linewidth]{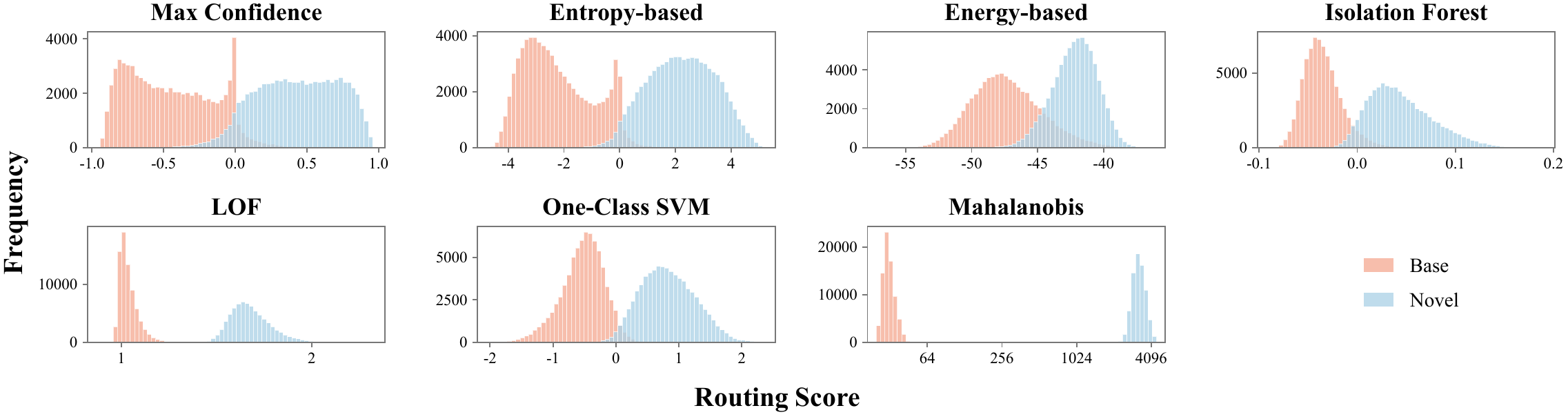}
    \caption{Routing-score distributions of seven routing methods on ImageNet.
    Mahalanobis produces a substantially larger gap between the Base and Novel distributions than the other methods.
    For clearer visualization, the horizontal axis of Mahalanobis is shown on a log$_2$ scale, while the others use linear scales.}
    \cn{
    七种路由方法在 ImageNet 上的分数分布。
    相比其他方法，Mahalanobis 在 Base 与 Novel 分布之间形成了明显更大的间隔。
    为便于观察，Mahalanobis 的横坐标采用 $\log_2$ 尺度，其余方法采用线性尺度。
    }
    \label{figure:Different Routing Methods}
}
\end{figure*}

\vspace{0.3\baselineskip}

\revision{
\subsection{Routing-Score Distribution Analysis}
\cn{路由分数分布分析}
}

\revision{
Following the likelihood-based motivation in Sec.~\ref{sec:petr}, we first examine the Mahalanobis-distance distributions on all 11 datasets.
The router measures how far the concatenated logits from the two prompt branches deviate from the Base training distribution, assigning lower distances to Base samples and higher distances to Novel samples.
As shown in Fig.~\ref{fig:mahalanobis_distributions}, the two distributions are clearly separated on ten datasets, whereas only a small tail overlap appears on EuroSAT.
These results provide dataset-level evidence for the strong discriminative ability of Mahalanobis routing and identify EuroSAT as its most challenging dataset.
}
\cn{
基于第~\ref{sec:petr} 节给出的似然解释，我们首先分析马氏距离在全部 11 个数据集上的分布。
该路由器衡量两组提示分支的拼接 logits 偏离 Base 训练分布的程度，因此 Base 样本通常具有较小距离，而 Novel 样本具有较大距离。
如图~\ref{fig:mahalanobis_distributions} 所示，两个分布在其中 10 个数据集上间隔明显，仅在 EuroSAT 上存在少量尾部重叠。
这些结果从各数据集的分布层面验证了马氏距离路由的区分能力，同时表明 EuroSAT 是最具挑战性的数据集。
}

\begin{table*}[!t]
\centering
\revision{
\caption{Comparison of different routing strategies on all 11 datasets. We report routing accuracy on Base and Novel samples, followed by Base, Novel, and harmonic-mean (HM) classification accuracy (\%). Average Logits combines the predictions of the two prompt branches without routing, while Oracle Routing uses the ground-truth Base/Novel membership.}
\cn{全部 11 个数据集上不同路由策略的比较。表中首先报告 Base 与 Novel 样本的路由准确率，随后报告 Base、Novel 及其调和平均（HM）分类准确率（\%）。Average Logits 不进行路由，而是融合两组提示分支的预测；Oracle Routing 则使用真实的 Base/Novel 归属进行路由。}
\small
\setlength{\tabcolsep}{3.65pt}
\begin{tabular}{l|cc|ccc|cc|ccc|cc|ccc}
\toprule
 & \multicolumn{5}{c|}{\textbf{Average}} & \multicolumn{5}{c|}{\textbf{ImageNet}} & \multicolumn{5}{c}{\textbf{Caltech101}} \\
\textbf{Method} & \multicolumn{2}{c|}{\textbf{Routing}} & \multicolumn{3}{c|}{\textbf{Classification}} & \multicolumn{2}{c|}{\textbf{Routing}} & \multicolumn{3}{c|}{\textbf{Classification}} & \multicolumn{2}{c|}{\textbf{Routing}} & \multicolumn{3}{c}{\textbf{Classification}} \\
 & \textbf{Base} & \textbf{Novel} & \textbf{Base} & \textbf{Novel} & \textbf{HM} & \textbf{Base} & \textbf{Novel} & \textbf{Base} & \textbf{Novel} & \textbf{HM} & \textbf{Base} & \textbf{Novel} & \textbf{Base} & \textbf{Novel} & \textbf{HM} \\
\midrule
Average Logits & -- & -- & 80.84 & 75.14 & 77.74 & -- & -- & 74.44 & 70.73 & 72.54 & -- & -- & 95.55 & 92.69 & 94.09 \\
Max Confidence & 87.40 & 84.87 & 84.30 & 78.62 & 81.14 & 94.81 & 93.83 & 83.59 & 74.83 & 78.97 & 97.85 & 95.71 & 97.91 & 95.74 & 96.81 \\
Entropy-based & 90.98 & 88.06 & 85.30 & 79.56 & 82.12 & 96.75 & 97.44 & 83.98 & 75.36 & 79.44 & 98.62 & 97.49 & 98.24 & 96.25 & 97.23 \\
Energy-based & 94.98 & 58.19 & 85.80 & 66.69 & 74.42 & 95.00 & 60.69 & 83.07 & 63.53 & 71.99 & 94.96 & 78.35 & 94.75 & 89.88 & 92.24 \\
Isolation Forest & 94.92 & 83.29 & 85.82 & 73.98 & 79.23 & 94.77 & 94.54 & 81.40 & 73.56 & 77.28 & 94.96 & 77.73 & 95.72 & 81.11 & 87.65 \\
LOF & 95.38 & 99.98 & 85.75 & 82.08 & 83.75 & 94.62 & \textbf{100.00} & 81.83 & \textbf{75.60} & 78.59 & 95.93 & 99.85 & 95.57 & 96.65 & 96.11 \\
One-Class SVM & 96.38 & 98.92 & 86.57 & 81.54 & 83.86 & 95.87 & 98.66 & 81.89 & 75.07 & 78.33 & 98.15 & \textbf{100.00} & 97.83 & \textbf{96.98} & 97.40 \\
\rowcolor{tabhighlight} Mahalanobis & \textbf{100.00} & \textbf{99.99} & \textbf{88.42} & \textbf{82.09} & \textbf{85.14} & \textbf{100.00} & \textbf{100.00} & \textbf{84.36} & \textbf{75.60} & \textbf{79.74} & \textbf{100.00} & \textbf{100.00} & \textbf{99.18} & \textbf{96.98} & \textbf{98.07} \\
\midrule
\textcolor{gray}{Oracle Routing} & \textcolor{gray}{100.00} & \textcolor{gray}{100.00} & \textcolor{gray}{88.42} & \textcolor{gray}{82.09} & \textcolor{gray}{85.14} & \textcolor{gray}{100.00} & \textcolor{gray}{100.00} & \textcolor{gray}{84.36} & \textcolor{gray}{75.60} & \textcolor{gray}{79.74} & \textcolor{gray}{100.00} & \textcolor{gray}{100.00} & \textcolor{gray}{99.18} & \textcolor{gray}{96.98} & \textcolor{gray}{98.07} \\
\bottomrule
\end{tabular}
\vspace{1em}\\
\begin{tabular}{l|cc|ccc|cc|ccc|cc|ccc}
\toprule
 & \multicolumn{5}{c|}{\textbf{OxfordPets}} & \multicolumn{5}{c|}{\textbf{StanfordCars}} & \multicolumn{5}{c}{\textbf{Flowers102}} \\
\textbf{Method} & \multicolumn{2}{c|}{\textbf{Routing}} & \multicolumn{3}{c|}{\textbf{Classification}} & \multicolumn{2}{c|}{\textbf{Routing}} & \multicolumn{3}{c|}{\textbf{Classification}} & \multicolumn{2}{c|}{\textbf{Routing}} & \multicolumn{3}{c}{\textbf{Classification}} \\
 & \textbf{Base} & \textbf{Novel} & \textbf{Base} & \textbf{Novel} & \textbf{HM} & \textbf{Base} & \textbf{Novel} & \textbf{Base} & \textbf{Novel} & \textbf{HM} & \textbf{Base} & \textbf{Novel} & \textbf{Base} & \textbf{Novel} & \textbf{HM} \\
\midrule
Average Logits & -- & -- & 91.09 & 91.89 & 91.47 & -- & -- & 76.15 & 77.53 & 76.83 & -- & -- & 92.12 & 74.70 & 82.48 \\
Max Confidence & 89.95 & 92.84 & 93.23 & 95.66 & 94.43 & 83.12 & 87.12 & 79.72 & 81.27 & 80.48 & 96.96 & 87.75 & 98.01 & 78.82 & 87.37 \\
Entropy-based & 92.22 & 93.85 & 93.99 & 96.14 & 95.05 & 92.56 & 92.71 & 82.23 & 82.99 & 82.61 & 97.72 & 90.66 & 98.29 & 80.31 & 88.39 \\
Energy-based & 95.00 & 64.11 & 93.71 & 86.33 & 89.75 & 94.98 & 62.00 & 82.50 & 67.48 & 74.17 & 94.97 & 70.47 & 95.35 & 66.86 & 78.50 \\
Isolation Forest & 95.00 & 97.00 & 94.84 & 96.74 & 95.76 & 95.19 & 62.94 & 82.91 & 63.37 & 71.81 & 94.97 & 99.74 & 95.66 & 83.48 & 89.15 \\
LOF & 95.71 & \textbf{100.00} & 94.08 & \textbf{98.23} & 96.11 & 95.24 & \textbf{100.00} & 82.39 & \textbf{84.51} & 83.44 & 96.14 & \textbf{100.00} & 96.11 & \textbf{83.69} & 89.47 \\
One-Class SVM & 98.33 & 99.87 & 95.89 & 98.36 & 97.11 & 95.11 & 96.50 & 82.44 & 82.14 & 82.29 & 97.12 & \textbf{100.00} & 97.56 & \textbf{83.69} & 90.09 \\
\rowcolor{tabhighlight} Mahalanobis & \textbf{100.00} & \textbf{100.00} & \textbf{96.79} & \textbf{98.23} & \textbf{97.50} & \textbf{100.00} & \textbf{100.00} & \textbf{84.22} & \textbf{84.51} & \textbf{84.36} & \textbf{100.00} & \textbf{100.00} & \textbf{99.46} & \textbf{83.69} & \textbf{90.90} \\
\midrule
\textcolor{gray}{Oracle Routing} & \textcolor{gray}{100.00} & \textcolor{gray}{100.00} & \textcolor{gray}{96.79} & \textcolor{gray}{98.23} & \textcolor{gray}{97.50} & \textcolor{gray}{100.00} & \textcolor{gray}{100.00} & \textcolor{gray}{84.22} & \textcolor{gray}{84.51} & \textcolor{gray}{84.36} & \textcolor{gray}{100.00} & \textcolor{gray}{100.00} & \textcolor{gray}{99.46} & \textcolor{gray}{83.69} & \textcolor{gray}{90.90} \\
\bottomrule
\end{tabular}
\vspace{1em}\\
\begin{tabular}{l|cc|ccc|cc|ccc|cc|ccc}
\toprule
 & \multicolumn{5}{c|}{\textbf{Food101}} & \multicolumn{5}{c|}{\textbf{FGVCAircraft}} & \multicolumn{5}{c}{\textbf{SUN397}} \\
\textbf{Method} & \multicolumn{2}{c|}{\textbf{Routing}} & \multicolumn{3}{c|}{\textbf{Classification}} & \multicolumn{2}{c|}{\textbf{Routing}} & \multicolumn{3}{c|}{\textbf{Classification}} & \multicolumn{2}{c|}{\textbf{Routing}} & \multicolumn{3}{c}{\textbf{Classification}} \\
 & \textbf{Base} & \textbf{Novel} & \textbf{Base} & \textbf{Novel} & \textbf{HM} & \textbf{Base} & \textbf{Novel} & \textbf{Base} & \textbf{Novel} & \textbf{HM} & \textbf{Base} & \textbf{Novel} & \textbf{Base} & \textbf{Novel} & \textbf{HM} \\
\midrule
Average Logits & -- & -- & 88.63 & 90.19 & 89.40 & -- & -- & 43.82 & 35.81 & 39.41 & -- & -- & 81.46 & 77.56 & 79.46 \\
Max Confidence & 94.11 & 95.58 & 90.75 & 92.69 & 91.70 & 55.58 & 49.47 & 40.74 & 36.39 & 38.17 & 90.50 & 85.08 & 83.53 & 79.80 & 81.62 \\
Entropy-based & 95.75 & 97.25 & 91.33 & 93.25 & 92.27 & 67.89 & 55.23 & 44.48 & 37.75 & 40.70 & 94.96 & 89.56 & 84.12 & 80.45 & 82.24 \\
Energy-based & 95.00 & 61.82 & 90.66 & 79.42 & 84.50 & 94.96 & 47.75 & 52.60 & 33.93 & 41.24 & 94.99 & 40.72 & 83.13 & 67.10 & 74.25 \\
Isolation Forest & 94.88 & 84.60 & 90.48 & 87.33 & 88.79 & 94.96 & 99.44 & 51.94 & 42.17 & 46.54 & 94.55 & 39.23 & 83.22 & 62.21 & 71.12 \\
LOF & 94.89 & \textbf{100.00} & 90.05 & \textbf{93.96} & 91.96 & 95.68 & \textbf{100.00} & 51.98 & \textbf{42.63} & 46.84 & 94.57 & \textbf{100.00} & 83.46 & \textbf{81.89} & 82.67 \\
One-Class SVM & 92.83 & 99.68 & 89.52 & 93.80 & 91.61 & 98.18 & 99.82 & 52.88 & \textbf{42.63} & 47.20 & 93.26 & 94.76 & 83.14 & 79.86 & 81.46 \\
\rowcolor{tabhighlight} Mahalanobis & \textbf{100.00} & \textbf{100.00} & \textbf{93.04} & \textbf{93.96} & \textbf{93.50} & \textbf{100.00} & \textbf{100.00} & \textbf{54.10} & \textbf{42.63} & \textbf{47.68} & \textbf{100.00} & \textbf{100.00} & \textbf{84.88} & \textbf{81.89} & \textbf{83.36} \\
\midrule
\textcolor{gray}{Oracle Routing} & \textcolor{gray}{100.00} & \textcolor{gray}{100.00} & \textcolor{gray}{93.04} & \textcolor{gray}{93.96} & \textcolor{gray}{93.50} & \textcolor{gray}{100.00} & \textcolor{gray}{100.00} & \textcolor{gray}{54.10} & \textcolor{gray}{42.63} & \textcolor{gray}{47.68} & \textcolor{gray}{100.00} & \textcolor{gray}{100.00} & \textcolor{gray}{84.88} & \textcolor{gray}{81.89} & \textcolor{gray}{83.36} \\
\bottomrule
\end{tabular}
\vspace{1em}\\
\begin{tabular}{l|cc|ccc|cc|ccc|cc|ccc}
\toprule
 & \multicolumn{5}{c|}{\textbf{DTD}} & \multicolumn{5}{c|}{\textbf{EuroSAT}} & \multicolumn{5}{c}{\textbf{UCF101}} \\
\textbf{Method} & \multicolumn{2}{c|}{\textbf{Routing}} & \multicolumn{3}{c|}{\textbf{Classification}} & \multicolumn{2}{c|}{\textbf{Routing}} & \multicolumn{3}{c|}{\textbf{Classification}} & \multicolumn{2}{c|}{\textbf{Routing}} & \multicolumn{3}{c}{\textbf{Classification}} \\
 & \textbf{Base} & \textbf{Novel} & \textbf{Base} & \textbf{Novel} & \textbf{HM} & \textbf{Base} & \textbf{Novel} & \textbf{Base} & \textbf{Novel} & \textbf{HM} & \textbf{Base} & \textbf{Novel} & \textbf{Base} & \textbf{Novel} & \textbf{HM} \\
\midrule
Average Logits & -- & -- & 77.43 & 63.29 & 69.61 & -- & -- & 85.37 & 76.90 & 80.87 & -- & -- & 83.13 & 75.21 & 78.97 \\
Max Confidence & 89.27 & 85.63 & 84.68 & 70.37 & 76.86 & 75.94 & 79.10 & 86.48 & 84.21 & 85.02 & 93.31 & 81.45 & 88.62 & 75.03 & 81.15 \\
Entropy-based & 91.05 & 86.84 & 85.53 & 70.77 & 77.45 & 77.76 & 83.66 & 86.94 & 85.93 & 86.02 & 95.47 & 84.01 & 89.16 & 75.97 & 81.95 \\
Energy-based & 94.91 & 46.62 & 85.65 & 52.42 & 64.88 & 95.00 & 63.86 & 94.50 & 75.61 & 83.56 & 94.98 & 43.70 & 87.87 & 50.98 & 63.54 \\
Isolation Forest & 94.91 & 94.57 & 85.96 & 68.52 & 76.22 & 94.91 & 78.12 & 94.32 & 80.56 & 86.57 & 94.98 & 88.26 & 87.61 & 74.78 & 80.60 \\
LOF & 95.79 & \textbf{100.00} & 85.65 & \textbf{72.83} & 78.72 & 94.76 & \textbf{99.96} & 94.34 & \textbf{89.10} & 91.65 & 95.90 & \textbf{100.00} & 87.78 & \textbf{83.76} & 85.72 \\
One-Class SVM & 95.52 & 99.52 & 85.42 & 72.14 & 78.22 & 98.10 & 99.35 & 96.17 & 88.67 & 92.26 & 97.76 & 99.93 & 89.50 & 83.63 & 86.47 \\
\rowcolor{tabhighlight} Mahalanobis & \textbf{100.00} & \textbf{100.00} & \textbf{88.39} & \textbf{72.83} & \textbf{79.86} & \textbf{99.98} & 99.85 & \textbf{97.55} & 89.07 & \textbf{93.12} & \textbf{100.00} & \textbf{100.00} & \textbf{90.68} & \textbf{83.76} & \textbf{87.08} \\
\midrule
\textcolor{gray}{Oracle Routing} & \textcolor{gray}{100.00} & \textcolor{gray}{100.00} & \textcolor{gray}{88.39} & \textcolor{gray}{72.83} & \textcolor{gray}{79.86} & \textcolor{gray}{100.00} & \textcolor{gray}{100.00} & \textcolor{gray}{97.55} & \textcolor{gray}{89.20} & \textcolor{gray}{93.19} & \textcolor{gray}{100.00} & \textcolor{gray}{100.00} & \textcolor{gray}{90.68} & \textcolor{gray}{83.76} & \textcolor{gray}{87.08} \\
\bottomrule
\end{tabular}
\label{table:RoutingComparison}
}
\end{table*}

\revision{
We then compare seven routing methods on EuroSAT, where Mahalanobis produces the weakest separation among the 11 datasets.
Max Confidence routes each sample to the prompt branch with the larger maximum softmax probability over all classes~\cite{max-confidence}, whereas Entropy-based routing selects the branch with lower predictive entropy.
Energy-based routing uses the energy score derived from the logits~\cite{energy-based}.
We also include three classical anomaly-detection methods: Isolation Forest~\cite{isolation-forest}, LOF~\cite{lof}, and One-Class SVM~\cite{one-class-SVM}.
As shown in Fig.~\ref{fig:eurosat_routing_distributions}, Max Confidence, Entropy-based, Energy-based, and Isolation Forest exhibit substantial overlap between Base and Novel samples.
LOF and One-Class SVM achieve clearer separation, while Mahalanobis provides the clearest and most balanced separation between the two groups.
}
\cn{
随后，我们在马氏距离区分效果最弱的 EuroSAT 上比较七种路由方法。
Max Confidence 将样本路由至最大类别 softmax 概率更高的提示分支 [max-confidence]，而 Entropy-based routing 选择预测熵更低的分支。
Energy-based routing 使用由 logits 计算得到的能量分数 [energy-based]。
我们还引入了三种经典异常检测方法：Isolation Forest [isolation-forest]、LOF [lof] 和 One-Class SVM [one-class-SVM]。
如图~\ref{fig:eurosat_routing_distributions} 所示，Max Confidence、Entropy-based、Energy-based 和 Isolation Forest 的 Base 与 Novel 分布存在明显重叠。
LOF 和 One-Class SVM 能够实现更清晰的分离，而 Mahalanobis 对两组样本的分离最为清晰且均衡。
}

\revision{
Because Mahalanobis, LOF, and One-Class SVM yield comparable classification performance on EuroSAT, we further compare them on ImageNet.
As shown in Fig.~\ref{figure:Different Routing Methods}, Mahalanobis produces a substantially larger gap between the Base and Novel distributions.
LOF and One-Class SVM also largely separate the two groups but with smaller margins, whereas Energy-based routing and Isolation Forest exhibit substantial overlap.
Max Confidence retains visible overlap near the decision boundary, while Energy-based routing and Isolation Forest exhibit more substantial overlap.
This comparison demonstrates that Mahalanobis provides more consistent separation across datasets.
}
\cn{
由于 Mahalanobis、LOF 和 One-Class SVM 在 EuroSAT 上取得了相近的分类性能，我们进一步在 ImageNet 上进行比较。
如图~\ref{figure:Different Routing Methods} 所示，Mahalanobis 在 Base 与 Novel 分布之间形成了明显更大的间隔。
LOF 和 One-Class SVM 也基本能够分离两组样本，但分布间隔相对较小，而 Energy-based routing 与 Isolation Forest 仍存在明显重叠。
Max Confidence 在决策边界附近仍存在可见重叠，而 Energy-based routing 与 Isolation Forest 的重叠更加明显。
这一比较表明，Mahalanobis 能够在不同数据集上提供更加稳定的分离效果。
}

\vfill

\revision{
\subsection{Comparison of Routing Methods}
\cn{路由方法比较}
}

\revision{
Table~\ref{table:RoutingComparison} evaluates whether the separation of routing-score distributions translates into improved classification performance.
LOF and One-Class SVM are the strongest alternative routers, consistent with their relatively clear distribution separation observed above.
Nevertheless, Mahalanobis achieves the highest average HM among all practical methods, matching Oracle Routing in average HM while consistently providing the best balance between Base and Novel classification across datasets.
These results demonstrate that its advantage extends beyond score-distribution separation to downstream classification performance.
}
\cn{
表~\ref{table:RoutingComparison} 进一步检验路由分数分布的分离能否转化为分类性能的提升。
LOF 和 One-Class SVM 是表现最强的替代路由方法，这与上文观察到的较清晰分布分离一致。
尽管如此，Mahalanobis 仍在所有实际可用的方法中取得最高的平均 HM，其平均 HM 与 Oracle Routing 持平，并在不同数据集上更稳定地平衡 Base 与 Novel 分类性能。
这表明马氏距离的优势不仅体现在路由分数分布的分离上，也能转化为下游分类性能的提升。
}

\vfill

\revision{
\subsection{Inference Efficiency}
Table~\ref{table:inference_efficiency} compares the inference efficiency of PETR with representative prompt learning and prompt ensembling methods under the same hardware and implementation settings.
Although this dual-branch design is more expensive than the single-branch PromptKD, PETR is consistently more efficient than BeyondSoleStrength across all three metrics.
These results demonstrate that PETR achieves branch specialization and conditional prompt selection with competitive efficiency among prompt-ensemble methods.
}
\cn{推理效率
表~\ref{table:inference_efficiency} 在相同硬件与实现设置下，将 PETR 的推理效率与代表性的提示学习及提示集成方法进行了比较。
PETR 需要计算两个专门化提示分支，因此其计算量高于单分支方法。训练无关路由器引入的开销可以忽略。
尽管该双分支设计的开销高于单分支 PromptKD，但 PETR 在三项效率指标上均优于 BeyondSoleStrength。
这些结果表明，PETR 在实现分支专门化与条件提示选择的同时，在提示集成方法中仍具有竞争力的推理效率。
}

\vfill

\begin{table}[H]
\centering
\revision{
    \caption{Inference-efficiency comparison under the same hardware and implementation settings; GFLOPs and latency are reported per image. Compared with PromptKD, PETR approximately doubles the inference cost. BeyondSoleStrength is less efficient than PETR across all three metrics.}
    \cn{
    相同硬件与实现设置下的推理效率比较；GFLOPs 和推理延迟均按单张图像统计。
    相比 PromptKD，PETR 的推理开销约为其两倍。BeyondSoleStrength 在三项效率指标上均不及 PETR。
    }
    \small
    \setlength{\tabcolsep}{3pt}
    \begin{tabular}{lccc}
    \toprule
        \textbf{Method} & \makecell{\textbf{GFLOPs}} & \makecell{\textbf{Latency (ms)}} & \textbf{FPS} \\
    \midrule
        PromptSRC~\cite{PromptSRC} & 174.04 & 1.50 & 702.63 \\
        PromptKD~\cite{PromptKD} & 34.40 & 1.34 & 747.48 \\
        BeyondSoleStrength~\cite{BeyondSoleStrength} & 74.17 & 2.81 & 354.96 \\
        \rowcolor{tabhighlight}  PETR (Ours) & 68.79 & 2.72 & 367.47 \\
    \midrule
        PETR Router & $4.7 \times 10^{-6}$ & 0.0042 & -- \\
    \bottomrule
    \end{tabular}
    \label{table:inference_efficiency}
}
\end{table}

\vfill

\begin{figure}[t]
    \centering
    \includegraphics[width=\columnwidth]{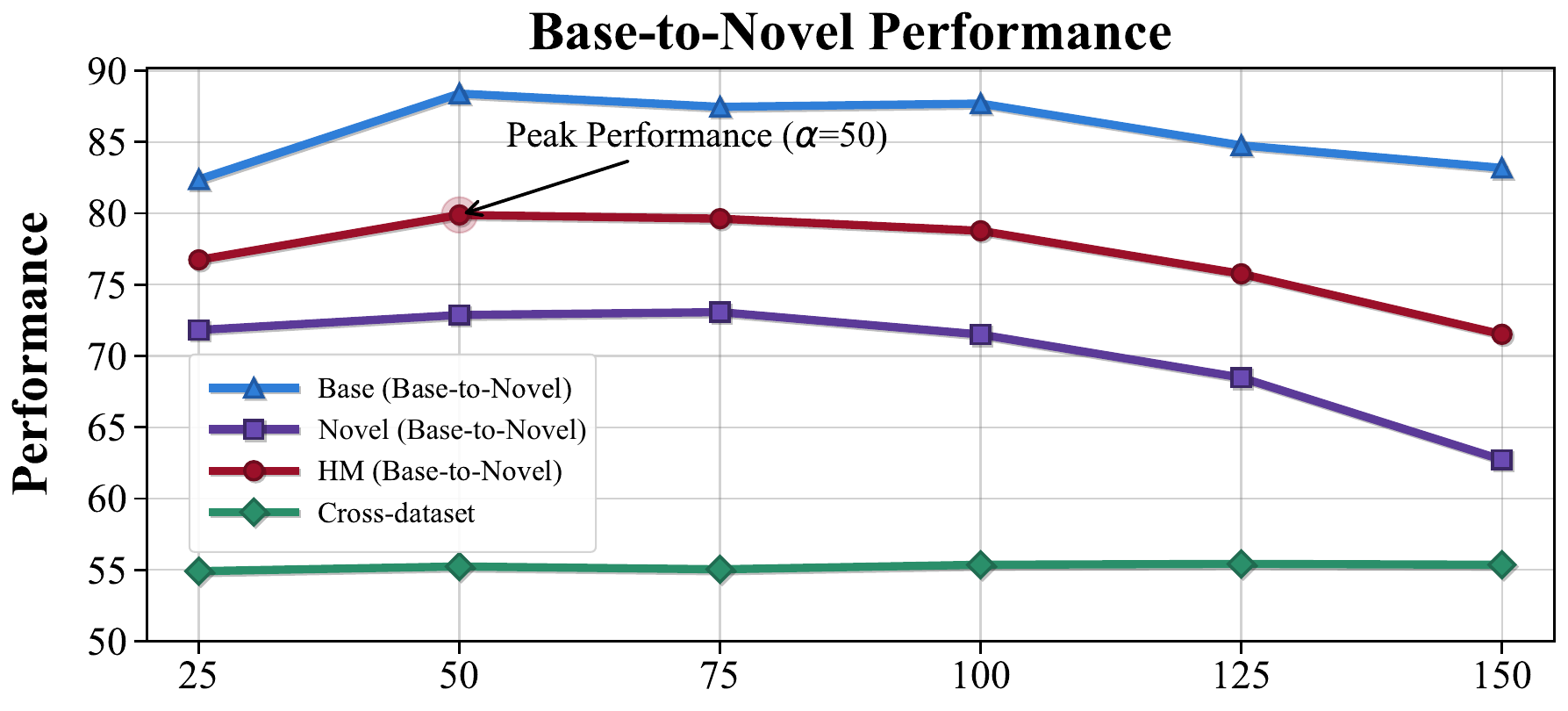}
    \caption{
    Effect of the distillation weight $\alpha$ on DTD.
\revision{
    Base-to-novel generalization results averaged over Base, Novel, and HM.
}
    Performance first increases and then decreases, peaking at $\alpha=50$.
    Cross-dataset evaluation results averaged across target datasets, showing minor sensitivity to $\alpha$.
    }
    \cn{
    蒸馏权重 α 对 DTD 性能的影响。
    （左）基础到新类别设置下的 Base / Novel / HM 结果：性能随 α 先升后降，并在 α=50 达到峰值。
    （右）跨数据集评估的平均结果：对 α 的敏感性较弱。
    }
    \label{figure:hyperparameter_comparison}
\end{figure}

\revision{
\subsection{Distillation Weight}
}
The distillation weight $\alpha$ controls the strength of the KD term in Eq.~\ref{equation:PETR + PromptKD}.
Figure~\ref{figure:hyperparameter_comparison} reports the sensitivity on DTD: under base-to-novel generalization, performance peaks at $\alpha=50$.
\revision{
For cross-dataset evaluation and domain generalization, we use the same $\alpha$ as in the corresponding base-to-novel setting.
}
\cn{
蒸馏权重
蒸馏权重 α 用于控制式~\ref{equation:PETR + PromptKD} 中 KD 项的强度。
如图~\ref{figure:hyperparameter_comparison} 所示，在 DTD 的基础到新类别设置下性能在 α=50 达到峰值。
对于跨数据集评估和领域泛化，我们在所有数据集中沿用基础到新类别设置下对应的 α。
}

\vfill

\revision{
\subsection{Threshold Scaling Hyperparameter}
The threshold scaling hyperparameter $\lambda$ in Eq.~\ref{equation:PETR} determines the threshold for routing samples to the Base or Novel prompt branch.
Since ImageNet has no separate validation split under this protocol, we consistently select $\lambda$ using the training splits of all 11 datasets without any test-set feedback.
Based on the average training-set routing accuracy across these datasets, we set $\lambda=1.4$ for all datasets.
}
\cn{阈值缩放超参数
式~\ref{equation:PETR} 中的阈值缩放超参数 λ 决定样本被路由至 Base 或 Novel 提示分支的判别阈值。
由于该实验协议中的 ImageNet 没有单独的验证集，我们统一使用 11 个数据集的训练划分选择 λ，不使用任何测试集反馈。
根据这些数据集训练集路由准确率的平均结果，我们在所有数据集上统一设置 λ=1.4。
}

\begin{table}[H]
    \caption{
\revision{
    Average training-set routing accuracies under different threshold scaling hyperparameters across 11 datasets.
    Base and Novel denote the proportions of corresponding training samples correctly routed to their respective prompt branches.
}
    The classification accuracy is highest when $\lambda = 1.4$.
    }
    \cn{
    不同阈值缩放超参数在 11 个数据集训练集上的平均路由准确率。
    Base 和 Novel 分别表示相应训练样本被正确路由至各自提示分支的比例。
    当 λ=1.4 时分类准确率最高。
    }
    \small \centering
    \begin{tabular}{lccc}
    \toprule
        \textbf{Classification Accuracy } & \multirow{2}{*}{\textbf{Base}} & \multirow{2}{*}{\textbf{Novel}} & \multirow{2}{*}{\textbf{Average}} \\
        \textbf{of Each Hyperparameter} \\ 
    \midrule
        $\lambda=0.6$ & 10.03 & 100.00 & 55.02 \\
        $\lambda=1.0$ & 98.52 & 100.00 & 99.26\\
        \rowcolor{tabhighlight}  $\lambda=1.4$  & 100.00 & 100.00 & \textbf{100.00}\\
        $\lambda=1.8$  & 100.00 & 99.98 & 99.99\\
    \bottomrule
    \end{tabular}
    \label{table:Hyperparameter}
\end{table}

\vfill

\section{Conclusion}

This paper proposes an ensembling framework, aiming to provide a general and effective solution to the trade-off between performance and generalization in prompt learning tasks. By leveraging the logits statistics of training samples from seen classes, the framework can dynamically determine whether a sample belongs to seen or unseen classes, and select the most suitable model output. Experimental results show that the framework significantly outperforms existing methods under various benchmark settings. This framework offers new insights for future research in multimodal learning. In future work, we plan to explore strategies to better distinguish fine-grained categories.
\cn{
本文提出一种提示集成框架，旨在为提示学习中“性能—泛化”权衡提供通用且有效的解决方案。
该框架利用已见类别训练样本的 logits 统计信息，动态判断输入更可能来自已见类别还是未见类别，并选择更合适的模型输出。
实验结果表明，该框架在多种基准设置下均显著优于现有方法。
该框架也为未来多模态学习研究提供了新的思路。
未来工作将探索更强的细粒度区分策略，以提升对细粒度类别的鲁棒性。
}

\clearpage


\bibliographystyle{IEEEtran}
\bibliography{refs}

\twocolumn[{%
\centering\LARGE\bfseries APPENDIX\par
\vspace{1em}
}]
\appendices
\section{Prompt Templates for PETR+CoOp}

For PETR+CoOp, we use the soft prompt trained by CoOp as $\theta_s$ and the original CLIP~\cite{CLIP} hard prompt as $\theta_u$. The hard prompt corresponds to a set of manually designed prompt templates for each dataset. Below, we list the specific prompt templates used for each dataset in our experiments.
\cn{
在 PETR+CoOp 中，我们将 CoOp 训练得到的 soft prompt 作为 $\theta_s$，原始 CLIP 的 hard prompt 作为 $\theta_u$。hard prompt 对应于每个数据集的一组手工设计的 prompt templates。下表列出了我们在各个数据集实验中使用的具体 prompt templates。
}

\begin{itemize}[label={},leftmargin=*]
\item \textbf{ImageNet~\cite{ImageNet}, ImageNetV2~\cite{ImageNet-V2}, ImageNetA~\cite{ImageNet-A}}
\begin{itemize}
    \item \texttt{"itap of a \{\}."}
    \item \texttt{"a bad photo of the \{\}."}
    \item \texttt{"a origami \{\}."}
    \item \texttt{"a photo of the large \{\}."}
    \item \texttt{"a \{\} in a video game."}
    \item \texttt{"art of the \{\}."}
    \item \texttt{"a photo of the small \{\}."}
\end{itemize}

\item \textbf{ImageNetR~\cite{ImageNet-R}}
\begin{itemize}
    \item \texttt{"the embroidered \{\}."}
    \item \texttt{"a drawing of a \{\}."}
    \item \texttt{"a cartoon \{\}."}
    \item \texttt{"a doodle of a \{\}."}
\end{itemize}

\item \textbf{ImageNetSketch~\cite{ImageNet-Sketch}}
\begin{itemize}
    \item \texttt{"a black and white drawing of a \{\}."}
    \item \texttt{"a sketch of a \{\}."}
\end{itemize}

\item \textbf{Caltech101~\cite{Caltech101}}
\begin{itemize}
    \item \texttt{"a photo of a \{\}."}
    \item \texttt{"a painting of a \{\}."}
    \item \texttt{"a plastic \{\}."}
    \item \texttt{"a sculpture of a \{\}."}
    \item \texttt{"a sketch of a \{\}."}
    \item \texttt{"a tattoo of a \{\}."}
    \item \texttt{"a toy \{\}."}
    \item \texttt{"a rendition of a \{\}."}
    \item \texttt{"a embroidered \{\}."}
    \item \texttt{"a cartoon \{\}."}
    \item \texttt{"a \{\} in a video game."}
    \item \texttt{"a plushie \{\}."}
    \item \texttt{"a origami \{\}."}
    \item \texttt{"art of a \{\}."}
    \item \texttt{"graffiti of a \{\}."}
    \item \texttt{"a drawing of a \{\}."}
    \item \texttt{"a doodle of a \{\}."}
    \item \texttt{"a photo of the \{\}."}
    \item \texttt{"a painting of the \{\}."}
    \item \texttt{"the plastic \{\}."}
    \item \texttt{"a sculpture of the \{\}."}
    \item \texttt{"a sketch of the \{\}."}
    \item \texttt{"a tattoo of the \{\}."}
    \item \texttt{"the toy \{\}."}
    \item \texttt{"a rendition of the \{\}."}
    \item \texttt{"the embroidered \{\}."}
    \item \texttt{"the cartoon \{\}."}
    \item \texttt{"the \{\} in a video game."}
    \item \texttt{"the plushie \{\}."}
    \item \texttt{"the origami \{\}."}
    \item \texttt{"art of the \{\}."}
    \item \texttt{"graffiti of the \{\}."}
    \item \texttt{"a drawing of the \{\}."}
    \item \texttt{"a doodle of the \{\}."}
\end{itemize}

\item \textbf{DescribableTextures~\cite{DTD}}
\begin{itemize}
    \item \texttt{"a photo of a \{\} texture."}
    \item \texttt{"a photo of a \{\} pattern."}
    \item \texttt{"a photo of a \{\} thing."}
    \item \texttt{"a photo of a \{\} object."}
    \item \texttt{"a photo of the \{\} texture."}
    \item \texttt{"a photo of the \{\} pattern."}
    \item \texttt{"a photo of the \{\} thing."}
    \item \texttt{"a photo of the \{\} object."}
\end{itemize}

\item \textbf{EuroSAT~\cite{EuroSAT}}
\begin{itemize}
    \item \texttt{"a centered satellite photo of \{\}."}
    \item \texttt{"a centered satellite photo of a \{\}."}
    \item \texttt{"a centered satellite photo of the \{\}."}
\end{itemize}

\item \textbf{FGVCAircraft~\cite{FGVCAircraft}}
\begin{itemize}
    \item \texttt{"a photo of a \{\}, a type of aircraft."}
    \item \texttt{"a photo of the \{\}, a type of aircraft."}
\end{itemize}

\item \textbf{OxfordFlowers~\cite{OxfordFlowers}}
\begin{itemize}
    \item \texttt{"a photo of a \{\}, a type of flower."}
\end{itemize}

\item \textbf{Food101~\cite{Food101}}
\begin{itemize}
    \item \texttt{"a photo of \{\}, a type of food."}
\end{itemize}

\item \textbf{OxfordPets~\cite{OxfordPets}}
\begin{itemize}
    \item \texttt{"a photo of a \{\}, a type of pet."}
\end{itemize}

\item \textbf{SUN397~\cite{SUN397}}
\begin{itemize}
    \item \texttt{"a photo of a \{\}."}
    \item \texttt{"a photo of the \{\}."}
\end{itemize}

\item \textbf{StanfordCars~\cite{StanfordCars}}
\begin{itemize}
    \item \texttt{"a photo of a \{\}."}
    \item \texttt{"a photo of the \{\}."}
    \item \texttt{"a photo of my \{\}."}
    \item \texttt{"i love my \{\}!"}
    \item \texttt{"a photo of my dirty \{\}."}
    \item \texttt{"a photo of my clean \{\}."}
    \item \texttt{"a photo of my new \{\}."}
    \item \texttt{"a photo of my old \{\}."}
\end{itemize}

\item \textbf{UCF101~\cite{UCF101}}
\begin{itemize}
    \item \texttt{"a photo of a person \{\}."}
    \item \texttt{"a video of a person \{\}."}
    \item \texttt{"a example of a person \{\}."}
    \item \texttt{"a demonstration of a person \{\}."}
\end{itemize}
\end{itemize}

\revision{
\section{Stability Across Random Seeds}
\cn{不同随机种子下的稳定性分析}
}

\revision{
To assess the stability of PETR, we repeat the complete training and evaluation procedure three times using different random seeds.
Table~\ref{tab:base2new_stats} reports the mean, sample standard deviation, and two-sided 95\% confidence interval of Base and Novel accuracy for each dataset.
The confidence intervals are computed from the three runs using the $t$-distribution with $n-1=2$ degrees of freedom.
}
\cn{
为评估 PETR 对随机因素的稳定性，我们使用三个不同的随机种子重复完整的训练与评估流程。
表~\ref{tab:base2new_stats} 给出了各数据集 Base 和 Novel 准确率的均值、样本标准差及双侧 95\% 置信区间。
置信区间根据三次运行的结果计算，因此采用自由度 n−1=2 的 t 分布。
}

\begin{table*}[!t]
\centering
\revision{
\caption{Base and Novel accuracies over three random seeds, reported as mean $\pm$ sample standard deviation (Std) with two-sided 95\% confidence intervals (CI). HM is omitted because, consistent with prior work and the main paper, it is computed from the mean Base and Novel accuracies rather than treated as a seed-wise statistic.}
\cn{
三个随机种子下的 Base 和 Novel 准确率，以均值、样本标准差及双侧 95\% 置信区间的形式报告。按照已有工作和论文正文的统计方式，HM 由 Base 与 Novel 的平均准确率计算，而不是作为逐随机种子统计量，因此本表不再报告 HM。
}
\label{tab:base2new_stats}
\setlength{\tabcolsep}{5pt}
\begin{tabular}{lcccc}
\toprule
\multirow{2}{*}{\textbf{Dataset}} & \multicolumn{2}{c}{\textbf{Base}} & \multicolumn{2}{c}{\textbf{Novel}} \\
\cmidrule(lr){2-3}\cmidrule(lr){4-5}
& \textbf{Mean $\pm$ Std} & \textbf{95\% CI} & \textbf{Mean $\pm$ Std} & \textbf{95\% CI} \\
\midrule
ImageNet & $84.36 \pm 0.05$ & $[84.23,\, 84.49]$ & $75.60 \pm 0.11$ & $[75.32,\, 75.88]$ \\
Caltech101 & $99.18 \pm 0.04$ & $[99.09,\, 99.28]$ & $96.98 \pm 0.32$ & $[96.20,\, 97.76]$ \\
OxfordPets & $96.79 \pm 0.19$ & $[96.33,\, 97.26]$ & $98.23 \pm 0.12$ & $[97.94,\, 98.52]$ \\
StanfordCars & $84.22 \pm 0.24$ & $[83.62,\, 84.82]$ & $84.51 \pm 0.06$ & $[84.37,\, 84.65]$ \\
OxfordFlowers & $99.46 \pm 0.05$ & $[99.33,\, 99.60]$ & $83.69 \pm 0.18$ & $[83.25,\, 84.13]$ \\
Food101 & $93.04 \pm 0.04$ & $[92.93,\, 93.14]$ & $93.96 \pm 0.12$ & $[93.65,\, 94.27]$ \\
FGVCAircraft & $54.10 \pm 0.33$ & $[53.28,\, 54.92]$ & $42.63 \pm 0.73$ & $[40.82,\, 44.44]$ \\
SUN397 & $84.88 \pm 0.11$ & $[84.61,\, 85.16]$ & $81.89 \pm 0.17$ & $[81.47,\, 82.32]$ \\
DescribableTextures & $88.39 \pm 0.40$ & $[87.39,\, 89.39]$ & $72.83 \pm 0.67$ & $[71.16,\, 74.50]$ \\
EuroSAT & $97.55 \pm 0.18$ & $[97.09,\, 98.00]$ & $89.07 \pm 1.93$ & $[84.27,\, 93.86]$ \\
UCF101 & $90.68 \pm 0.33$ & $[89.85,\, 91.51]$ & $83.76 \pm 0.11$ & $[83.48,\, 84.04]$ \\
\bottomrule
\end{tabular}
}
\end{table*}

\revision{
PETR exhibits consistently low variation across random seeds.
The standard deviations range from 0.04 to 0.40 for Base, while those for Novel remain at or below 0.73 on all datasets except EuroSAT (1.93).
Across the 11 datasets in Table~II of the main paper, PETR improves the dataset-wise HM over PromptKD by 1.31\% on average.
Treating datasets as paired units, a two-sided exact Wilcoxon signed-rank test confirms that this cross-dataset improvement is statistically significant ($W=0$, $p=0.00098$).
These results demonstrate that PETR is generally stable across random runs and datasets, although its Novel accuracy on EuroSAT exhibits relatively larger variation.
}
\cn{
PETR 在不同随机种子下总体表现出较小的性能波动。
Base 准确率的标准差为 0.04～0.40；除 EuroSAT 为 1.93 外，其余数据集的 Novel 准确率标准差均不超过 0.73。
在正文 Table II 的 11 个数据集上，PETR 相比 PromptKD 的逐数据集 HM 平均提升 1.31\%。
以数据集为配对单位的双侧精确 Wilcoxon 符号秩检验表明，该跨数据集提升具有统计显著性（W=0，p=0.00098）。
这些结果表明 PETR 在不同随机运行和数据集上总体稳定，但其在 EuroSAT 上的 Novel 准确率存在相对更大的波动。
}

\revision{
\section{Data Splits and Prompt-Branch Utilization}
\cn{数据划分与提示分支使用比例}
}

\revision{
PETR follows the dataset splits used by PromptKD~\cite{PromptKD}.
Within the Train split, labeled Base images are used to learn the Base-oriented prompt, unlabeled Novel images are used to learn the Novel-oriented prompt, and Base images are used to estimate the router statistics.
Neither test subset is involved in prompt learning, router-statistics estimation, or routing-threshold selection.
}
\cn{
PETR 沿用 PromptKD [PromptKD] 的数据集划分。
在 Train 划分中，带标注的 Base 图像用于学习面向 Base 的提示，无标注的 Novel 图像用于学习面向 Novel 的提示，Base 图像则用于估计路由统计量。
两个测试子集均不参与提示学习、路由统计量估计或路由阈值选择。
}

\begin{table}[!t]
\centering
\revision{
\caption{Dataset split sizes and inferred prompt-branch utilization under the PromptKD protocol adopted by PETR. Train denotes the images available during training, while Test Base and Test Novel are disjoint evaluation subsets. The last two columns report the proportions of test samples routed to the Base-oriented and Novel-oriented prompt branches, respectively.}
\cn{
PETR 所采用的 PromptKD 协议下的数据集划分规模及推算得到的提示分支使用比例。
Train 表示训练阶段可用的图像，Test Base 和 Test Novel 是彼此独立的评估子集。
最后两列分别表示测试样本被路由至面向 Base 和面向 Novel 提示分支的比例。
}
\label{tab:promptkd_data_split}
\small
\setlength{\tabcolsep}{3.20pt}
\begin{tabular}{lccccc}
\toprule
\textbf{Dataset} & \textbf{Train} & \textbf{Test Base} & \textbf{Test Novel} & \textbf{Base} & \textbf{Novel} \\
\midrule
ImageNet & 1,281,167 & 25,000 & 25,000 & 50.0\% & 50.0\% \\
Caltech101 & 4,128 & 1,549 & 916 & 62.8\% & 37.2\% \\
OxfordPets & 2,944 & 1,881 & 1,788 & 51.3\% & 48.7\% \\
StanfordCars & 6,509 & 4,002 & 4,039 & 49.8\% & 50.2\% \\
Flowers102 & 4,093 & 1,053 & 1,410 & 42.8\% & 57.3\% \\
Food101 & 50,500 & 15,300 & 15,000 & 50.5\% & 49.5\% \\
FGVCAircraft & 3,334 & 1,666 & 1,667 & 50.0\% & 50.0\% \\
SUN397 & 15,880 & 9,950 & 9,900 & 50.1\% & 49.9\% \\
DTD & 2,820 & 864 & 828 & 51.1\% & 48.9\% \\
EuroSAT & 13,500 & 4,200 & 3,900 & 51.9\% & 48.2\% \\
UCF101 & 7,639 & 1,934 & 1,849 & 51.1\% & 48.9\% \\
\bottomrule
\end{tabular}
}
\end{table}

\revision{
The prompt-branch utilization is inferred from the Base/Novel test-set sizes in Table~\ref{tab:promptkd_data_split} and the corresponding routing accuracies reported in Table~VI of the main paper.
Since Mahalanobis routing is perfect on ten datasets and nearly perfect on EuroSAT, the utilization of the two prompt branches is almost identical to the corresponding Base/Novel test proportions.
These results confirm that both specialized prompt branches are substantially utilized rather than the router collapsing to a single branch.
}
\cn{
提示分支的使用比例根据表~\ref{tab:promptkd_data_split} 中 Base/Novel 测试集的规模以及正文表 VI 报告的相应路由准确率推算得到。
由于马氏距离路由在十个数据集上完全正确，并在 EuroSAT 上接近完全正确，因此两个提示分支的使用比例几乎与相应的 Base/Novel 测试样本占比一致。
这些结果表明，两组专门化提示均得到充分使用，路由器没有退化为仅选择单一分支。
}

\begin{table*}[ht]
\caption{Performance of our proposed PETR+CoOp method on multiple datasets, compared with CoOp. The table reports results for base classes, novel classes, and their harmonic mean. $\Delta$ denotes the performance improvement of our methods over the compared methods.}
\cn{
表1. 我们提出的PETR+CoOp方法在多个数据集上的性能表现，并与CoOp进行了对比。表中列出了基础类别、新类别以及它们的调和平均值的结果。Δ表示我们的方法相较于对比方法的性能提升。
}
    \small \centering
    \begin{tabular}{ll|rrr|rrr|rrr}
    \toprule
    \multicolumn{2}{c|}{\multirow{2}[2]{*}{}} & \multicolumn{3}{c|}{\textbf{Average}} & \multicolumn{3}{c|}{\textbf{ImageNet}} & \multicolumn{3}{c}{\textbf{Caltech101}}  \\
    \multicolumn{2}{c|}{} & \multicolumn{1}{c}{\textbf{Base}} & \multicolumn{1}{c}{\textbf{Novel}} & \multicolumn{1}{c|}{\textbf{HM}} & \multicolumn{1}{c}{\textbf{Base}} & \multicolumn{1}{c}{\textbf{Novel}} & \multicolumn{1}{c|}{\textbf{HM}} & \multicolumn{1}{c}{\textbf{Base}} & \multicolumn{1}{c}{\textbf{Novel}} & \multicolumn{1}{c}{\textbf{HM}} \\
    \midrule
    \multicolumn{2}{l|}{CoOp~\cite{CoOp}} & \textbf{82.69}  & 63.22  & 71.66  & \textbf{76.47}  & 67.88  & 71.92  & 98.00  & 89.81  & 93.73 \\
    \rowcolor{tabhighlight} \multicolumn{2}{l|}{PETR+CoOp (ours)} & 82.39 & \textbf{73.82} & \textbf{77.49} & 76.45 & \textbf{70.67} & \textbf{73.45} & \textbf{98.06} & \textbf{93.67} & \textbf{95.81} \\
    \multicolumn{2}{l|}{$\Delta$} & \decre{-0.30} & \incre{+10.60} & \incre{+5.83} & \decre{-0.02} & \incre{+2.79} & \incre{+1.53} & \incre{+0.06} & \incre{+3.86} & \incre{+2.08} \\
    \bottomrule
    \end{tabular}
    \vspace{0.6em}\\
    \begin{tabular}{ll|rrr|rrr|rrr}
    \toprule
    \multicolumn{2}{c|}{\multirow{2}[2]{*}{}} & \multicolumn{3}{c|}{\textbf{OxfordPets}} & \multicolumn{3}{c|}{\textbf{StanfordCars}} & \multicolumn{3}{c}{\textbf{Flowers102}}  \\
    \multicolumn{2}{c|}{} & \multicolumn{1}{c}{\textbf{Base}} & \multicolumn{1}{c}{\textbf{Novel}} & \multicolumn{1}{c|}{\textbf{HM}} & \multicolumn{1}{c}{\textbf{Base}} & \multicolumn{1}{c}{\textbf{Novel}} & \multicolumn{1}{c|}{\textbf{HM}} & \multicolumn{1}{c}{\textbf{Base}} & \multicolumn{1}{c}{\textbf{Novel}} & \multicolumn{1}{c}{\textbf{HM}} \\
    \midrule
    \multicolumn{2}{l|}{CoOp~\cite{CoOp}} & 93.67 & 95.29 & 94.47 & \textbf{78.12} & 60.40 & 68.13 & 97.60 & 59.67 & 74.06 \\
    \rowcolor{tabhighlight} \multicolumn{2}{l|}{PETR+CoOp (ours)} & \textbf{94.05} & \textbf{97.09} & \textbf{95.55} & 77.54 & \textbf{74.20} & \textbf{75.83} & \textbf{97.78} & \textbf{77.38} & \textbf{86.39} \\
    \multicolumn{2}{l|}{$\Delta$} & \incre{+0.38} & \incre{+1.80} & \incre{+1.08} & \decre{-0.58} & \incre{+13.80} & \incre{+7.70} & \incre{+0.18} & \incre{+17.71} & \incre{+12.33} \\
    \bottomrule
    \end{tabular}
    \vspace{0.6em}\\
    \begin{tabular}{ll|rrr|rrr|rrr}
    \toprule
    \multicolumn{2}{c|}{\multirow{2}[2]{*}{}} & \multicolumn{3}{c|}{\textbf{Food101}} & \multicolumn{3}{c|}{\textbf{FGVCAircraft}} & \multicolumn{3}{c}{\textbf{SUN397}}  \\
    \multicolumn{2}{c|}{} & \multicolumn{1}{c}{\textbf{Base}} & \multicolumn{1}{c}{\textbf{Novel}} & \multicolumn{1}{c|}{\textbf{HM}} & \multicolumn{1}{c}{\textbf{Base}} & \multicolumn{1}{c}{\textbf{Novel}} & \multicolumn{1}{c|}{\textbf{HM}} & \multicolumn{1}{c}{\textbf{Base}} & \multicolumn{1}{c}{\textbf{Novel}} & \multicolumn{1}{c}{\textbf{HM}} \\
    \midrule
    \multicolumn{2}{l|}{CoOp~\cite{CoOp}} & \textbf{88.33} & 82.26 & 85.19 & \textbf{40.44} & 22.30 & 28.75 & 80.60 & 65.89 & 72.51 \\
    \rowcolor{tabhighlight} \multicolumn{2}{l|}{PETR+CoOp (ours)} & 88.01 & \textbf{91.06} & \textbf{89.51} & 39.02 & \textbf{34.81} & \textbf{36.79} & \textbf{80.63} & \textbf{77.42} & \textbf{78.99} \\
    \multicolumn{2}{l|}{$\Delta$} & \decre{-0.32} & \incre{+8.80} & \incre{+4.32} & \decre{-1.42} & \incre{+12.51} & \incre{+8.04} & \incre{+0.03} & \incre{+11.53} & \incre{+6.48} \\
    \bottomrule
    \end{tabular}
    \vspace{0.6em}\\
    \begin{tabular}{ll|rrr|rrr|rrr}
    \toprule
    \multicolumn{2}{c|}{\multirow{2}[2]{*}{}} & \multicolumn{3}{c|}{\textbf{DTD}} & \multicolumn{3}{c|}{\textbf{EuroSAT}} & \multicolumn{3}{c}{\textbf{UCF101}}  \\
    \multicolumn{2}{c|}{} & \multicolumn{1}{c}{\textbf{Base}} & \multicolumn{1}{c}{\textbf{Novel}} & \multicolumn{1}{c|}{\textbf{HM}} & \multicolumn{1}{c}{\textbf{Base}} & \multicolumn{1}{c}{\textbf{Novel}} & \multicolumn{1}{c|}{\textbf{HM}} & \multicolumn{1}{c}{\textbf{Base}} & \multicolumn{1}{c}{\textbf{Novel}} & \multicolumn{1}{c}{\textbf{HM}} \\
    \midrule
    \multicolumn{2}{l|}{CoOp~\cite{CoOp}} & 79.44 & 41.18 & 54.24 & \textbf{92.19} & 54.74 & 68.69 & \textbf{84.69} & 56.05 & 67.46 \\
    \rowcolor{tabhighlight} \multicolumn{2}{l|}{PETR+CoOp (ours)} & \textbf{80.40} & \textbf{59.02} & \textbf{68.07} & 91.09 & \textbf{58.98} & \textbf{71.60} & 84.50 & \textbf{77.77} & \textbf{81.00} \\
    \multicolumn{2}{l|}{$\Delta$} & \incre{+0.96} & \incre{+17.84} & \incre{+13.83} & \decre{-1.10} & \incre{+4.24} & \incre{+2.91} & \decre{-0.19} & \incre{+21.72} & \incre{+13.54} \\
    \bottomrule
    \end{tabular}
\label{table:CoOp Base2Novel}
\end{table*}

\begin{table*}[t]
    \caption{Cross-dataset benchmark evaluation. PETR+CoOp's performance outperforms CoOp. PETR+CoOp outperforms others on 8 out of 10 datasets, and attains a higher average performance.}
    \cn{跨数据集评估结果。PETR+CoOp 的性能超过了 CoOp。与之前的方法相比，我们的方法在 10 个数据集中的 8 个上表现更好，平均值也高于之前的方法。}
    \small \centering
    \begin{tabular}{lccccccccccc}
        \hline\noalign{\smallskip}
        \textbf{Method} & \rotatebox{90}{\textbf{Caltech101}} & \rotatebox{90}{\textbf{OxfordPets}} & \rotatebox{90}{\textbf{StandfordCars}} & \rotatebox{90}{\textbf{Flowers102}} & \rotatebox{90}{\textbf{Food101}} & \rotatebox{90}{\textbf{FGVCAircraft}} & \rotatebox{90}{\textbf{SUN397}} & \rotatebox{90}{\textbf{DTD}} & \rotatebox{90}{\textbf{EuroSAT}} & \rotatebox{90}{\textbf{UCF101}} & \rotatebox{90}{\textbf{Avg.}} \\
        \hline\noalign{\smallskip}
        CoOp~\cite{CoOp} & \textbf{93.70} & \textbf{89.14} & 64.51 & 68.71 & 85.30 & 18.47 & 64.15 & 41.92 & 46.39 & 66.55 & 63.88 \\
        \rowcolor{tabhighlight} PETR+CoOp (ours) & 92.37 & 89.10 & \textbf{66.27} & \textbf{70.77} & \textbf{85.90} & \textbf{24.18} & \textbf{65.35} & \textbf{45.51} & \textbf{48.83} & \textbf{68.68} & \textbf{65.70} \\
        \hline\noalign{\smallskip}
    \end{tabular}
    \label{table:CoOp Cross-dataset evaluation} 
\end{table*}

\section{PETR CoOp Performance Detail}

\noindent \textbf{Base-to-novel Generalization.}
Table~\ref{table:CoOp Base2Novel} presents the detailed results of PETR+CoOp under the base-to-novel experimental setting across multiple datasets. Compared to CoOp, PETR+CoOp achieves nearly identical performance on base classes, indicating that the router introduces minimal loss to the classification of seen categories. For novel classes, PETR+CoOp achieves a substantial improvement, with performance close to that of CLIP with prompt ensembling and significantly higher than CoOp. As a result, the harmonic mean of base and novel class performance is much higher than CoOp.
\cn{
表~\ref{table:CoOp Base2Novel} 展示了 PETR+CoOp 在 base-to-novel 实验设置下，多个数据集上的详细实验结果。与 CoOp 相比，PETR+CoOp 在 base 类别上的性能几乎一致，说明 router 对已见类别的分类影响极小。在新类别上，PETR+CoOp 的性能有显著提升，接近于 CLIP 加上 prompt ensembling 的效果，远高于 CoOp。因此，base 和 novel 类别的调和平均（HM）大幅超过 CoOp。
}

These results demonstrate that PETR+CoOp can greatly enhance the generalization ability of CoOp to novel categories without any additional training. The improvement is mainly attributed to the effective combination of the CoOp-trained soft prompt and the original CLIP hard prompt, allowing the model to leverage the strengths of both. Overall, PETR+CoOp provides a simple yet effective way to boost the generalization of prompt learning methods.
\cn{
这些结果表明，PETR+CoOp 能在无需额外训练的前提下，大幅提升 CoOp 对新类别的泛化能力。这一提升主要得益于对 CoOp 训练得到的 soft prompt 与原始 CLIP hard prompt 的有效结合，使模型能够同时利用两者的优势。总体来看，PETR+CoOp 为提升 prompt learning 方法的泛化性提供了一种简单而有效的方案。
}

\noindent \textbf{Cross-dataset Evaluation.}
Table~\ref{table:CoOp Cross-dataset evaluation} presents the cross-dataset evaluation results of PETR+CoOp compared with CoOp and other baseline methods. PETR+CoOp achieves the best performance on 8 out of 10 datasets and obtains a higher average performance overall.The slightly lower performance on Caltech101 datasets may be due to the high visual similarity between certain classes (such as Leopards vs. Cougar face in Caltech101), which increases the difficulty of cross-dataset generalization. These results demonstrate the strong generalization ability of PETR+CoOp in cross-dataset scenarios.
\cn{
表~\ref{table:CoOp Cross-dataset evaluation} 展示了 PETR+CoOp 与 CoOp 及其他基线方法在跨数据集评测下的实验结果。PETR+CoOp 在 10 个数据集中的 8 个上取得了最佳性能，并且整体平均性能更高。这 Caltech101 数据集上略低的表现，可能是由于某些类别之间存在较高的视觉相似性（例如 FGVCAircraft 中的 Boeing 737-300 与 Boeing 737-400，或 Caltech101 中的 Leopards 与 Cougar face），从而增加了跨数据集泛化的难度。这些结果表明，PETR+CoOp 在跨数据集场景下具有很强的泛化能力。
}

\noindent \textbf{Domain Generalization.}
Table~\ref{table:CoOp Domain generalization} evaluates the direct transferability of models trained on ImageNet to a range of out-of-domain datasets. On the target datasets, PETR surpasses previous methods on 3 out of 4 datasets and achieves a higher average accuracy overall. The slightly lower performance on the ImageNet V2 dataset may be attributed to its subtle distributional differences from the original ImageNet, which are challenging for models to capture.

\begin{table}[t]
    \caption{Domain generalization results. PETR+CoOp outperforms previous methods on 3 out of 4 datasets, and achieves a higher average accuracy.}
    \cn{领域泛化结果。}
    \small \centering
    \begin{tabular}{lcccccc}
        \hline\noalign{\smallskip}
        \textbf{Method} & \textbf{-V2} & \textbf{-Sketch} & \textbf{-A} & \textbf{-R} & \textbf{Avg.} \\
        \hline\noalign{\smallskip}
        CoOp & \textbf{64.20} & 47.99  & 49.71  & 75.21  & 59.28\\
        \rowcolor{tabhighlight} PETR+CoOp & 62.30 & \textbf{48.00} & \textbf{50.70} & \textbf{77.50} & \textbf{59.63} \\
    \end{tabular}
    \label{table:CoOp Domain generalization}
\end{table}

\end{document}